\documentclass[10pt,twocolumn,letterpaper]{article}
\usepackage[pagenumbers]{cvpr}

\usepackage{iftex}
\ifPDFTeX
  \usepackage[accsupp]{axessibility}
\fi
\ifXeTeX
  \usepackage{newunicodechar}
  \newunicodechar{’}{\textquoteright}
  \newunicodechar{“}{\textquotedblleft}
  \newunicodechar{”}{\textquotedblright}
  \newunicodechar{—}{\textemdash}
\fi

\definecolor{blue4}{RGB}{85,142,213}
\definecolor{red4}{RGB}{255,102,102}

\usepackage{colortbl}
\usepackage{multirow}
\usepackage{listings}
\usepackage{pifont}
\usepackage{adjustbox}

\ProvideDocumentEnvironment{nolinenumbers}{}{}{}

\usepackage{tcolorbox}
\tcbuselibrary{skins,breakable}

\newcommand{\cmark}{\ding{51}}
\newcommand{\xmark}{\ding{55}}

\newtcolorbox{graylist}{
  enhanced,
  breakable,              %
  colback=gray!5,
  colframe=gray!15,
  boxrule=0pt,
  arc=0mm,
  left=1mm, right=1mm, top=1mm, bottom=1mm,
  width=\linewidth,       %
  coltext=black!90,
  fontupper=\slshape,
  before upper={\raggedright\setlength{\parskip}{2pt}},
}

\definecolor{cvprblue}{rgb}{0.21,0.49,0.74}
\usepackage[pagebackref,breaklinks,colorlinks,allcolors=cvprblue]{hyperref}

\title{\texorpdfstring{\includegraphics[height=0.65cm]{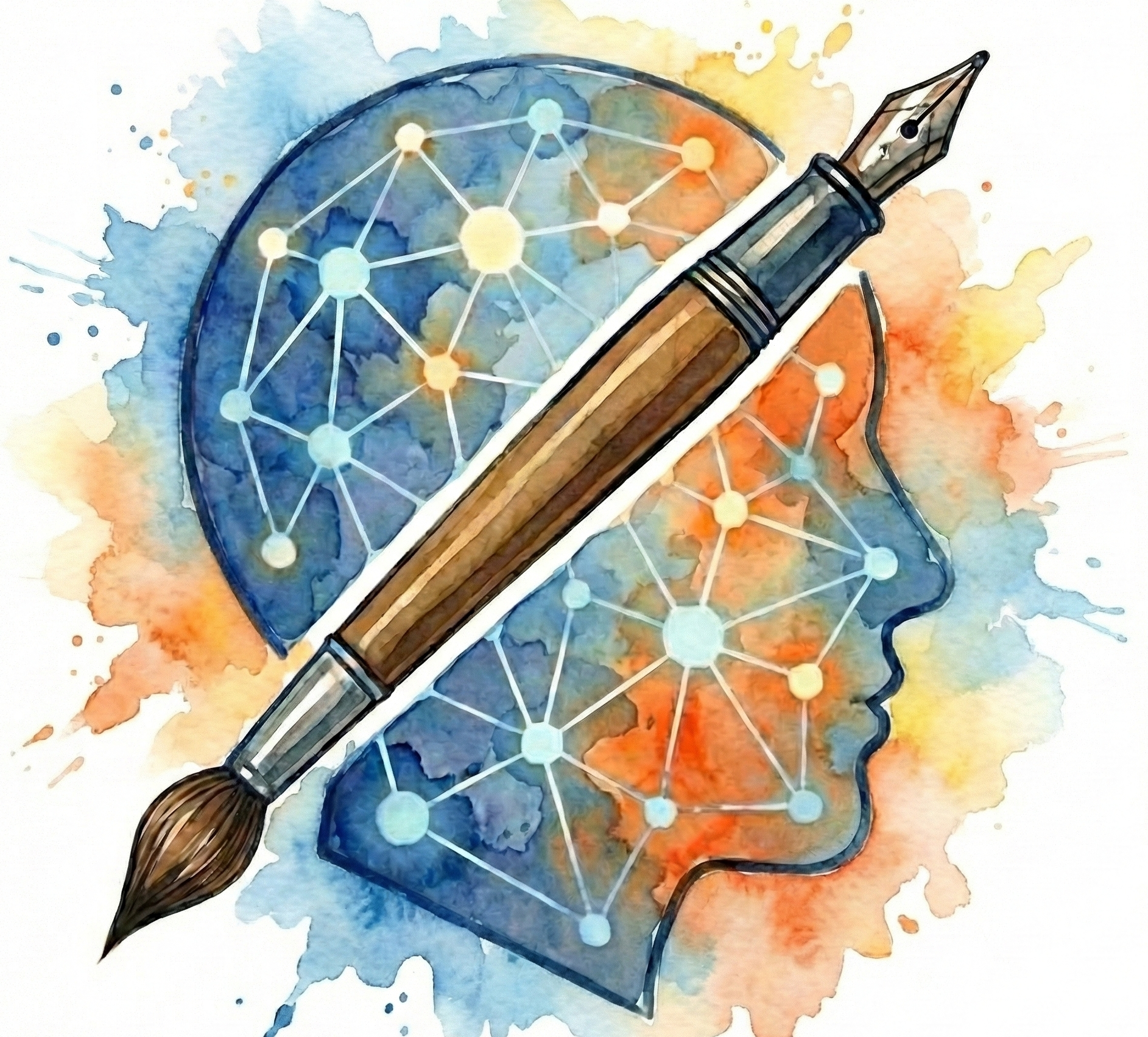} }{}EditThinker: Enhancing Image Editors via Iterative Reasoning with a Plug-and-Play VLM}
\hypersetup{pdftitle={EditThinker: Enhancing Image Editors via Iterative Reasoning with a Plug-and-Play VLM},pdfauthor={}}

\makeatletter
\def\@fnsymbol#1{\ensuremath{\ifcase#1\or \dagger\or \ddagger\or
   \mathsection\or \mathparagraph\or \|\or **\or \dagger\dagger
   \or \ddagger\ddagger \else\@ctrerr\fi}}
\makeatother

\author{\small
\begin{tabular}{c}
\textbf{Hongyu Li}$^{1,2}$
\quad
\textbf{Manyuan Zhang}$^{2}$\thanks{Project Leader.}
\quad
\textbf{Dian Zheng}$^{2,3}$
\quad
\textbf{Ziyu Guo}$^{2, 4}$
\quad
\textbf{Yimeng Jia}$^{2}$ \\[1ex]
\textbf{Kaituo Feng}$^{2,3}$
\quad
\textbf{Hao Yu}$^{5}$
\quad
\textbf{Yexin Liu}$^{2}$
\quad
\textbf{Yan Feng}$^{2}$
\quad
\textbf{Peng Pei}$^{2}$ \\[1ex]
\textbf{Xunliang Cai}$^{2}$
\quad
\textbf{Linjiang Huang}$^{1}$
\quad
\textbf{Hongsheng Li}$^{3}$
\quad
\textbf{Si Liu}$^{1}$\thanks{Corresponding Author.} \\[1ex]
\normalfont 
$^1$Beihang University \quad 
$^2$Meituan \quad 
$^3$CUHK MMLab \quad 
$^4$CUHK IMIXR \quad 
$^5$Tsinghua University \\[1ex]
{ \normalfont 
\texttt{Project Page: \!\!\!\!\!\url{https://appletea233.github.io/think-while-edit}}} \\
\end{tabular}
}

\hypersetup{pdfauthor={Hongyu Li, Manyuan Zhang, Dian Zheng, Ziyu Guo, Yimeng Jia, Kaituo Feng, Hao Yu, Yexin Liu, Yan Feng, Peng Pei, Xunliang Cai, Linjiang Huang, Hongsheng Li, Si Liu}}

\begin{document}
\addtocontents{toc}{\protect\setcounter{tocdepth}{-1}}
\maketitle
\begin{abstract}
Image editors can produce realistic images while still failing to satisfy the requested changes. Improving their instruction following requires identifying these failures and translating visual feedback into instructions that the editor can execute. We introduce \textbf{EditThinker}, an 8B vision-language model (VLM) that enhances existing image editors through iterative reasoning while keeping their weights fixed. EditThinker jointly critiques semantic consistency and visual quality, reasons about editing failures, and refines the instruction in a structured response. It interacts with the editor through a \textit{Critique-Refine-Repeat} loop, providing a plug-and-play interface for feedback-driven editing. Learning effective refinements requires more than imitating plausible expert suggestions: the instructions must work with the capabilities of the underlying editor. We therefore combine supervised fine-tuning on expert demonstrations with reinforcement learning from executed editing outcomes, using format, critic, and edit rewards. To support this training, we construct {\upshape\textsc{ThinkEdit}}-167k, a dataset of 167k step-wise samples with unified supervision for critique, reasoning, and instruction refinement. Across four tested editors, EditThinker improves overall scores on GEdit-Bench-EN, ImgEdit-Bench, RISE-Bench, and KRIS-Bench. Ablations examine the joint model design, data allocation, and training across VLM backbones. In our GEdit round-budget study, refinement achieves higher scores with fewer editor calls than best-of-$N$ selection. We will release our data construction framework, datasets, and models.
\end{abstract}

\section{Introduction}

Instruction-based image editing modifies a source image according to natural language instructions while preserving content that should remain unchanged. Recent editors built on strong generative models~\citep{qwen-image,flux-kontext,omnigen2,team2025longcat,sd3-medium} produce realistic images, yet can still misinterpret requested changes or alter unrelated regions. Addressing these failures requires understanding both the intended modification and the actual editing result. A reasoning module must translate visual feedback into instructions the editor can execute.

Existing strategies enhance instructions~\citep{wang2025promptenhancer} or score candidate edits~\citep{luo2025editscore}. We focus on learning to revise instructions from visual feedback while keeping the editor fixed. This setting lets the thinker guide different editors through their instruction interfaces.

The central challenge is to make reasoning useful for execution. A VLM may correctly describe an editing error yet suggest a revision that the same editor cannot realize. Because the editor is fixed, the thinker must learn which changes to the instruction lead to better visual outcomes. We therefore connect two learning objectives: understanding the current edit and producing a refinement that works when executed. This motivates a joint Critic-Refiner design and a training procedure that combines expert demonstrations with feedback from actual edits.

We introduce \textbf{EditThinker}, an 8B VLM trained to jointly critique editing results and refine instructions. Given the source image, original instruction, previous instruction, and current result, it generates a reasoning trace, semantic-consistency and perceptual-quality scores, and a revised instruction. Critique and refinement share a single structured response, with reasoning generated before the revised instruction. This design connects assessment of the visual result to the next editing action. We use a \textbf{Critique-Refine-Repeat} loop to execute these actions: the fixed editor applies the revised instruction to the source image, and the thinker inspects the new result. The loop stops at a score threshold or iteration budget.

Training this module requires both demonstrations of useful reasoning and feedback about whether its instructions work in practice. We collect expert-guided instruction-refinement trajectories with existing editors, pairing editing states with reasoning traces and revised instructions. An additional scorer evaluates intermediate results; filtering and truncation select trajectories that match or improve on the initial edit, from which we extract \textbf{\textsc{ThinkEdit}-167k} step-wise samples. We partition the data by training objective, using stable or consistently high-quality samples for supervised fine-tuning (SFT) and trajectories with high score variance for reinforcement learning (RL). The resulting 140k SFT and 27k RL samples connect source images, editing states, reasoning traces, and refined instructions. This division separates learning stable refinement patterns from optimizing decisions on trajectories with varying outcomes. Random and reversed data allocations provide alternative configurations for examining this choice across the two training stages.

SFT initializes the thinker with the expert's reasoning and structured output format. RL then updates the thinker from actual editor executions, using rewards for format adherence, critique, and editing improvement. This second stage addresses the gap between imitating an expert's instruction and producing one that works with a fixed editor. We study the resulting quality-iteration trade-off: different reward configurations can favor editing quality or fewer editor calls, and combining rewards offers a balance in our experiments. The editor remains fixed throughout this training and refinement process. At inference, the trained 8B thinker provides both critique and instruction refinement, interacting with existing editors through \mbox{their image-and-text interface.}

We evaluate EditThinker on four benchmarks covering general and reasoning-based image editing. With EditThinker, all four editors achieve higher overall scores on GEdit-Bench-EN, ImgEdit-Bench, RISE-Bench, and KRIS-Bench. Ablations examine rewards, iteration budgets, role coupling, reasoning order, and the SFT/RL data partition. Paired comparisons of additional VLM backbones before and after EditThinker training assess the applicability of the training procedure. The GEdit round-budget study also shows fewer editor calls than \mbox{best-of-$N$ selection.}

The main contributions of this work are:
\begin{itemize}[leftmargin=*,nosep]
    \interlinepenalty=10000\relax
    \item \textbf{Plug-and-Play EditThinker Design:} Our Critique-Refine-Repeat loop uses visual critiques to revise instructions, which a fixed editor applies to the source image.
    \item \textbf{Dataset and Two-Stage Training:} We construct \textsc{ThinkEdit}-167k with step-wise refinement samples and train with expert demonstrations for SFT and feedback from executed edits for RL.
    \item \textbf{Superior Performance and Ablations:} EditThinker improves overall scores for all four editors across four benchmarks. We ablate model design and data allocation, compare VLM backbones, and study \mbox{editing quality and editor calls.}
\end{itemize}
We will release our data construction framework, \mbox{datasets, and models.}

\section{Related work}
\label{sec:related_work}

\subsection{Image editing}
Instruction-based editing builds on diffusion and flow-based image generation~\citep{song2020score,rombach2022high,lipman2022flow,liu2025cot,yan2025gpt}. Early approaches used image-guided denoising and inversion~\citep{meng2021sdedit,hertz2022prompt,mokady2023null}, or explicit spatial conditioning~\citep{zhang2023adding,ye2023ip}, to control edits. InstructPix2Pix~\citep{brooks2023instructpix2pix} introduced natural-language control through instruction tuning. Subsequent work has developed dedicated instruction-based editors~\citep{zhang2025context,step1x} and unified multimodal models~\citep{qwen-image,openai_image_api,omnigen2,uniworldv2}, with FLUX.1 Kontext using flow matching for in-context editing~\citep{flux-kontext}. Preserving unedited content while satisfying complex instructions remains a challenge for single-pass editors.

\subsection{Iterative reasoning and refinement}
\label{sec:iterative_refinement}
Self-Refine~\citep{madaan2023selfrefine} revises outputs with self-generated feedback. Reflexion~\citep{shinn2023reflexion} stores verbal reflections for later trials. Idea2Img~\citep{yang2023idea2img} revises prompts with GPT-4V, and GenArtist~\citep{wang2024genartist} decomposes requests, selects tools, and verifies results. Reflect-DiT~\citep{li2025reflectdit} trains a VLM judge and adapts a diffusion transformer to use its feedback. ReflectionFlow~\citep{zhuo2025reflectionflow} fine-tunes a visual reflector and diffusion corrector. ReasonEdit~\citep{yin2025reasonedit} adapts its reasoner and generator separately, then jointly fine-tunes them. MIRA~\citep{zeng2026mira} trains a plug-and-play VLM with SFT and GRPO. Its editor applies each predicted atomic edit to the latest image. EditThinker jointly critiques the result, predicts semantic and perceptual quality scores, and rewrites the full instruction. The fixed editor applies each rewrite to the original image.

\subsection{Reward models for image editing}
Research on multimodal models as judges~\citep{chen2024mllm,zhang2025upme} motivates model-based rewards for visual generation~\citep{xu2023imagereward,niu2025wise}. Prior approaches use discrete scores or logit-based values~\citep{gong2025onereward,wu2024q}. Specialized methods adapt feedback to image editing: EditReward~\citep{wu2025editreward} trains an evaluator using human preferences, and EditScore~\citep{luo2025editscore} develops reward models for online RL. Scalar rewards evaluate generation outcomes~\citep{zhang2024large}. Multimodal planning provides a way to choose the next action~\citep{liu2025cot,yang2024mastering}. EditThinker learns from expert scores and executed edits during training. At inference, its reasoning trace connects the critique of the current image to \mbox{the next instruction.}

\section{\textbf{EditThinker}}
\label{sec:methodology}

\begin{figure*}[t]
    \centering 
    \includegraphics[width=\textwidth]{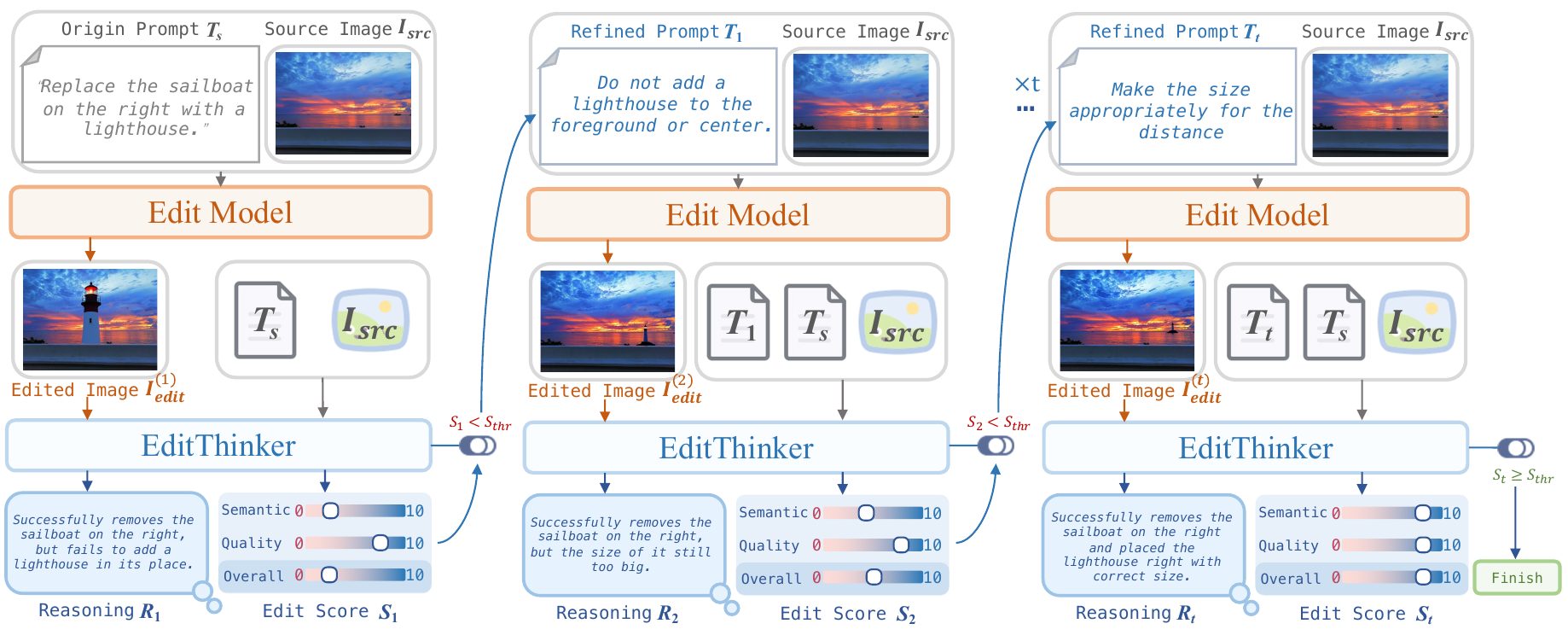} 
    \caption{\textbf{Critique-Refine-Repeat pipeline.} The editor first applies the original instruction $T_s$ to the source image $I_{src}$. Given this initial edit, EditThinker generates score $S_t$, reasoning $R_t$, and refined instruction $T_t$. The fixed editor applies $T_t$ to $I_{src}$ until the stopping threshold or iteration budget is reached.}
    \label{fig:fig2} 
\end{figure*}

EditThinker is an 8B VLM trained to critique editing results and refine instructions in a \textit{Critique-Refine-Repeat} loop while keeping the underlying image editor fixed.

\subsection{Overall framework}

A single-pass editor takes a source image $I_{src}$ and the original instruction $T_s$ to produce an edited image. EditThinker adds feedback to this process by inspecting the result and refining the instruction for a subsequent editing round.

The workflow separates a VLM-based Thinker for critique and instruction refinement from an Editor for execution (\eg, Qwen-Image-Edit or FLUX.1 Kontext). We train the Thinker via SFT and RL, and use the Editor through its image-and-text interface without updating its parameters. At iteration $t$, the Thinker inspects the previous output $I_{edit}^{t-1}$ and generates reasoning $R_t$, semantic and quality scores $S_t$, and a refined instruction $T_t$ in one structured response:

\begin{equation} 
    (S_t, R_t, T_t) = \text{Thinker}(I_{src}, I_{edit}^{t-1}, T_{t-1}, T_s).
\label{eq:cycle}
\end{equation}

Then the Editor executes the new instruction $T_t$ on the source image $I_{src}$, generating the updated result $I_{edit}^t$ as:

\begin{equation} \label{eq:editor}
    I_{edit}^t = \text{Editor}(I_{src}, T_t).
\end{equation}

We stop when the geometric mean of the predicted semantic and quality scores reaches $S_{\mathrm{thr}}=8$, or after $N=5$ editing rounds by default.

\subsection{Design of the EditThinker}
\label{sec:task_formulate}

We formulate EditThinker as a \textbf{dual-role model} that jointly evaluates and plans. Its Critic and Refiner share the same VLM and produce one structured response, allowing instruction refinement to use the critique of the \mbox{current editing result.}

We order the output so that the model reasons about the previous edit and generates its scores before producing the refined instruction $T_t$. Generating $R_t$ before $T_t$ lets the instruction condition on the visual critique of $I_{src}$ and $I_{edit}^{t-1}$. We evaluate both role coupling and output order in \cref{tab:additional_design_choices}.

To implement this dual-role design, we define a structured input-output format that explicitly encodes the evaluation-then-planning process.

\noindent\textbf{Input Tuple.} 
EditThinker receives a multimodal tuple ($I_{src}$, $I_{edit}^{t-1}$, $T_s$, $T_{t-1}$) at each iteration $t$, providing complete context of the editing state: $I_{src}$ and $T_s$ represents the original reference, $I_{edit}^{t-1}$ is the current result to be critiqued, and $T_{t-1}$ is the previous instruction that produced it.

\noindent\textbf{Structured Output Format.} 
The output is a structured text string that serializes EditThinker's reasoning process:

\vspace{0.3em}
\begin{center}
    \noindent\fbox{\parbox{\dimexpr\linewidth-2\fboxsep-2\fboxrule}{%
        \centering
        \texttt{<think>} Reasoning process... \texttt{</think>} \\
        \texttt{<score>} [$S_{sem}$, $S_{qual}$] \texttt{</score>} \\
        \texttt{<answer>} Refined prompt $T_t$ \texttt{</answer>}%
    }}
\end{center}
\vspace{0.3em}

Here, $S_{qual}$ measures the perceptual quality of $I_{edit}^{t-1}$. Given $I_{src}$, $S_{sem}$ measures the edit's adherence to $T_s$.

\subsection{Training of EditThinker}

Training EditThinker to perform this dual-role task requires a specialized dataset and a multi-stage training strategy. We adopt a two-stage approach: first, supervised fine-tuning (SFT) to learn the output format and basic reasoning, followed by reinforcement learning (RL) to optimize instruction refinement based on actual editing feedback. The data construction process is detailed in \cref{sec:data_construction}.

\subsubsection{Supervised fine-tuning (cold start)}
\label{sec:sft}
Using the expert (GPT-4.1) demonstration dataset (detailed in { \cref{sec:data_construction}}), the base VLM learns to adopt our structured I/O format (\eg, $<$think$>$, $<$score$>$, $<$answer$>$), mimic the expert's reasoning style, and understand the principles of critiquing and refining instructions.

\subsubsection{Reinforcement learning tuning (RLT)}
\label{sec:rl}
SFT imitates expert reasoning but may produce instructions that fail under the practical limitations of specific editors. We therefore optimize EditThinker with GRPO (Group Relative Policy Optimization) using actual editing feedback and a multi-component reward for both roles (\cref{sec:task_formulate}).

\noindent\textbf{Critic Reward.}
We use GPT-4.1~\citep{openai2025gpt41} as the critic expert. For each dimension $d\in\{\mathrm{sem},\mathrm{qual}\}$, we compare the Thinker's predicted score $S_{d,t}$ with the expert score $\mathcal{E}_d$ for the same edit $I_{edit}^{t-1}$. The critic reward is the negative mean absolute error:
\begin{equation} 
R_{\text{critic}} =
-\frac{1}{2}\sum_{d\in\{\mathrm{sem},\mathrm{qual}\}}
\left|S_{d,t}-\mathcal{E}_d(I_{src},I_{edit}^{t-1},T_s)\right|.
\end{equation}

Both the Thinker and the expert assess $I_{edit}^{t-1}$, before executing the new instruction $T_t$. This reward trains the critique of the current result; the edit reward below evaluates the subsequent editing improvement.

\noindent\textbf{Edit Reward.} 
To train instruction refinement, we reward gains between successive edits as scored by the same expert.
\begin{equation}
R_{\text{edit}} = \mathcal{E}(I_{src}, I_{edit}^{t}, T_s) - \mathcal{E}(I_{src}, I_{edit}^{t-1}, T_s)
\end{equation}
The reward is positive only when $T_t$ produces a better edit, grounding refinement in execution feedback.

The final reward $R_{\text{total}}$ is as follows:
\begin{equation}
R_{overall} = \alpha  R_{format} + \beta R_{critic} + \gamma R_{edit}
\end{equation}
where $R_{format}$ is the reasoning format reward and $(\alpha,\beta,\gamma)=(0.5,0.1,0.4)$. Design ablations appear in \cref{tab:additional_design_choices}; reward-weight ablations are in Appendix~\ref*{sec:additional_reward_weights}.

\section{\textsc{ThinkEdit} dataset}
\label{sec:data_construction}

\begin{figure*}[t]
    \centering 
    \includegraphics[width=\textwidth]{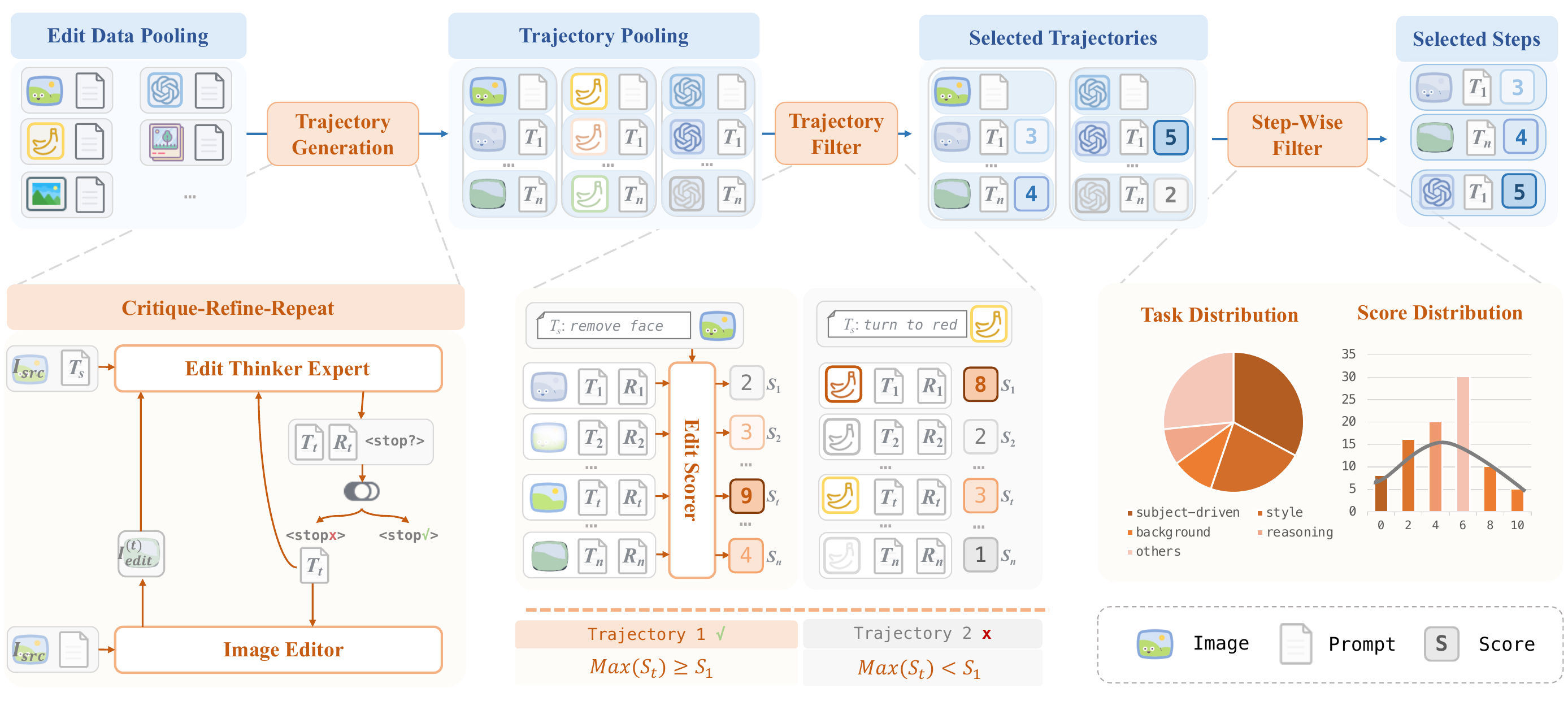} 
    \caption{\textbf{ThinkEdit data construction.} Expert-guided editing generates trajectories, which are scored, filtered for improvement, and truncated at their best step. We extract step-wise samples, balance task and score distributions, and partition them for SFT and RL.}
    \label{fig:fig3} 
\end{figure*}

To train the {EditThinker}, we require a high-quality dataset that captures the multi-turn \textit{Critique-Refine-Repeat } cycle. As shown in \cref{fig:fig3}, we designed an automated data construction pipeline to simulate this process, consisting of four sequential steps: Trajectory Generation, Trajectory Filter, Step-wise Filter, and Data Partition. This pipeline allowed us to construct our \textbf{\textsc{ThinkEdit}-167k} dataset.

\subsection{Trajectory generation}

The first stage focuses on simulating the multi-turn ``Critique-Refine-Repeat'' cycle. The pipeline begins with an Edit Data Pool containing diverse ($I_{src}$, $T_s$) pairs. At each step $t$, the edit thinker expert (GPT-4.1) evaluates the current state (based on $I_{src}$, $T_s$, and $I_{edit}^{t-1}$) and outputs instruction $T_t$, reasoning trace $R_t$, and a \mbox{$\langle\text{stop}\rangle$ token.}

Notably, the expert does not output a score ($S_t$). Instead, it directly determines when to halt the process by issuing a \mbox{$\langle\text{stop}\rangle$ token.} A separate scorer assigns per-step scores for trajectory filtering.

If a $\langle\text{stop}\rangle$ token is not issued, the image editor uses the new $T_t$ to produce $I_{edit}^{t}$. This loop continues until the expert triggers the $\langle\text{stop}\rangle$ condition (or a max-iteration limit $N$ is hit), thus completing a full trajectory.

\subsection{Trajectory filter}

Since the edit thinker expert only generates refined instructions and a $\langle\text{stop}\rangle$ token without quality scores, we employ an additional edit scorer to evaluate each step $I_{edit}^{(t)}$ and assign a score $S_t$. After scoring all steps ($S_1, \dots, S_n$), we apply a two-stage filtering process:

\noindent\textbf{Filter Failed Trajectories.} 
We retain only trajectories where at least one subsequent step ($t>1$) achieves a score higher than or equal to the initial step (i.e., $\max(S_{t>1}) \ge S_1$). Trajectories failing this condition are discarded.

\noindent\textbf{Truncate Kept Trajectories.} 
For retained trajectories, we identify the step $k$ with the highest score ($S_k = \max(S_{t \ge 1})$) and truncate the trajectory to include only steps from 1 to $k$. All subsequent steps ($t > k$) are discarded.

\subsection{Step-wise filter}
Finally, we process the curated trajectories from the Trajectory Filter to create the final training data through two steps:

\noindent\textbf{Sample Extraction.} First, we unroll the truncated trajectories. Each individual step $t$ within a trajectory is converted into a distinct training sample. This sample pairs an input tuple ($I_{src}$, $I_{edit}^{t-1}$, $T_s$, $T_{t-1}$) with its corresponding ground-truth expert output ($R_t, T_t$), retaining the score $S_t$ as metadata for subsequent filtering.

\noindent\textbf{Expert Demonstrations.}
The expert compares each intermediate image with the original request and checks whether details outside the requested edit remain intact. Its critique identifies errors such as missing objects, incorrect placement, or unwanted changes. The revised instruction retains edits that worked and specifies how to address the remaining errors. A misplaced object can call for a more precise spatial relation. Lost subject details can call for an explicit preservation constraint. See Fig.~\ref*{fig:expert_prompt} of the supplement for the expert prompt.

\noindent\textbf{Distribution Balancing.} 
We filter samples to balance task types (e.g., object removal, color modification, and addition) and score levels, ensuring coverage of both editing tasks and output quality.

\subsection{SFT and RL data split}

\noindent\textbf{Data Sources.}
We sample 40K editing instances each from OpenGPT-4o-Image and ShareGPT-4o-Image, and 60K from Pico-Banana-400K. GPT-4.1 guides trajectory generation with OmniGen2, FLUX.1-Kontext [Dev], and Qwen-Image-Edit as the editors. Trajectory filtering retains 70K trajectories from the 140K raw instances.

We partition the curated data by training objective. For SFT, we select stable, high-quality examples. For RL, we select challenging trajectories with room for improvement.

\noindent\textbf{RL Dataset.}
We select 10k trajectories with high score variance ($\text{Var}(S_t)>\theta$, $\theta=3.0$), balancing task and score distributions. These challenging trajectories provide informative improvement signals and yield 27k \mbox{step-wise RL samples.}

\noindent\textbf{SFT Dataset.}
From the remaining 60K trajectories, we select 140k step-wise samples with low score variance or consistently high quality, providing reliable demonstrations of stable instruction refinement.

\section{Experiments}
\subsection{Experimental setup}
\noindent\textbf{Training.}
EditThinker uses Qwen3-VL-8B-Instruct~\citep{bai2025qwen3}. We train on \textsc{ThinkEdit}-SFT-140k for one epoch with learning rate $2\times10^{-5}$ and batch size 32, then on \textsc{ThinkEdit}-RL-27k for one epoch with learning rate $2\times10^{-6}$, global batch size 16, GRPO rollout group size 8, and KL coefficient $10^{-3}$. MAX\_PIXELS is $1024\times1024$. Training uses 8 H800 GPUs for approximately 48 hours.

\noindent\textbf{Inference.}
We keep OmniGen2~\citep{omnigen2}, FLUX.1-Kontext [Dev]~\citep{flux-kontext}, Qwen-Image-Edit~\citep{qwen-image}, and Longcat-Image-Edit~\citep{team2025longcat} fixed. We use at most $N=5$ editing rounds and stop early when the geometric mean of the semantic and quality scores reaches 8. See Appendix~\ref*{sec:additional_stopping} for hyperparameters and the stopping rule.

\noindent\textbf{Benchmarks and Baselines.}
We evaluate general editing on GEdit-Bench-EN~\citep{step1x} and ImgEdit-Bench~\citep{imgedit}, and reasoning-based editing on RISE-Bench~\citep{rise} and KRIS-Bench~\citep{kris}.

\noindent\textbf{Evaluation Protocol.}
The reproduced single-pass baseline applies the user's instruction once to the source image. Its EditThinker counterpart starts with the same editor checkpoint, source image, and instruction. In each subsequent round, the fixed editor applies the revised instruction to the source image. Each round counts one editor call, including the initial edit. For RISE-Bench, we report the overall success rate and the corresponding rates for temporal, causal, spatial, and logical edits.

\twocolumn[{\begin{minipage}{\textwidth}
\captionsetup{type=table}
\caption{Image editing benchmark results. $^{\ddag}$ denotes reproduced single-pass baselines with official checkpoints. Bold marks the highest score within each pair. $\Delta$ shows the change from the listed baseline. See Appendix~\ref*{app:kris} for KRIS-Bench details.}
\centering
\label{tab:Img-Bench_all}
\definecolor{etdeltagreen}{RGB}{28,112,62}
\definecolor{etdeltared}{RGB}{166,46,46}
\newcommand{\etgain}[1]{\textcolor{etdeltagreen}{+#1}}
\newcommand{\etloss}[1]{\textcolor{etdeltared}{$-$#1}}
\renewcommand{\arraystretch}{0.90}
\resizebox{\textwidth}{!}{%
\begin{tabular}{l|ccc|c|ccccc|c}
\toprule
\multirow{2}{*}{\textbf{Model}}
& \multicolumn{3}{c|}{\textbf{GEdit-Bench-EN}}
& \multicolumn{1}{c|}{\textbf{ImgEdit-Bench}}
& \multicolumn{5}{c|}{\textbf{RISE-Bench}}
& \textbf{KRIS-Bench} \\
\cmidrule(lr){2-4}\cmidrule(lr){5-5}\cmidrule(lr){6-10}\cmidrule(lr){11-11}
& \textbf{G\_SC} & \textbf{G\_PQ} & \textbf{G\_O} & \textbf{Overall}
& \textbf{Temporal} & \textbf{Causal} & \textbf{Spatial} & \textbf{Logical}
& \textbf{Overall} & \textbf{Overall} \\
\midrule
\rowcolor{blue4!20}
\multicolumn{11}{c}{\textit{Open-source Models}} \\
OmniGen~\citep{xiao2025omnigen} & 5.96 & 5.89 & 5.06 & 2.96 & 1.2 & 1.0 & 0.0 & 1.2 & 0.8 & 28.85 \\
Step1X-Edit~\citep{step1x} & 7.09 & 6.76 & 6.70 & 3.06 & 0.0 & 2.2 & 2.0 & 3.5 & 1.9 & 43.29 \\
ICEdit~\citep{zhang2025context} & 4.94 & 7.39 & 4.87 & 3.05 & - & - & - & - & - & 40.70 \\
HiDream-E1~\citep{hidream} & 5.66 & 6.06 & 5.01 & 3.17 & 0.0 & 0.0 & 0.0 & 0.0 & 0.0 & 44.72 \\
BAGEL~\citep{bagel} & 7.36 & 6.83 & 6.52 & 3.20 & 2.4 & 5.6 & 14.0 & 1.2 & 6.1 & 56.40 \\
BAGEL (w/ CoT)~\citep{bagel} & 7.70 & 6.51 & 6.66 & 3.56 & 5.9 & 17.8 & 21.0 & 1.2 & 11.9 & 60.41 \\
OmniGen2~\citep{omnigen2} & 7.16 & 6.77 & 6.41 & 3.44 & 0.0 & 2.2 & 7.0 & 2.3 & 3.0 & 49.71 \\
Ovis-U1~\citep{wang2025ovis} & - & - & 6.42 & 4.00 & 1.2 & 3.3 & 4.0 & 2.4 & 2.8 & - \\
FLUX.1-Kontext [Dev]~\citep{flux-kontext} & 6.52 & 7.38 & 6.00 & 3.52 & 2.3 & 5.5 & 13.0 & 1.2 & 5.8 & 49.54 \\
UniWorld-V1~\citep{lin2025uniworld} & 4.93 & 7.43 & 4.85 & 3.26 & - & - & - & - & 2.22 & 50.27 \\
Step1X-Edit v1.1~\citep{step1x} & 7.66 & 7.35 & 6.97 & 3.90 & - & - & - & - & - & 51.59 \\
Qwen-Image-Edit~\citep{qwen-image} & 8.00 & 7.86 & 7.56 & 4.27 & 4.7 & 10.0 & 17.0 & 2.4 & 8.9 & - \\
Qwen-Image-Edit-2509~\citep{qwen2025edit2509} & 8.15 & 7.86 & 7.54 & 4.35 & - & - & - & - & - & 63.71 \\
Qwen-Image-Edit-2511~\citep{qwen2025edit2511} & 8.30 & 8.20 & 7.88 & 4.51 & 21.2 & 18.9 & 31.0 & 4.7 & 19.4 & 62.04 \\
Uni-CoT~\citep{qin2025uni} & 7.91 & 6.24 & 6.74 & 3.65 & 8.2 & 18.9 & 20.0 & 1.2 & 12.5 & 68.00 \\
UniWorld-V2~\citep{uniworldv2} & 8.39 & 8.02 & 7.83 & 4.49 & - & - & - & - & - & - \\
FireRed-Image-Edit-1.0~\citep{team2026firered} & 8.36 & 8.25 & 7.94 & 4.56 & - & - & - & - & - & - \\
JoyAI-Image-Edit~\citep{song2026joyaiimage} & 8.83 & 8.12 & 8.28 & 4.46 & - & - & - & - & - & 63.44 \\
\midrule
\rowcolor{red4!20}
\multicolumn{11}{c}{\textit{Proprietary Models}} \\
Gemini 2.0 Flash Image~\citep{kampf2025experiment} & 6.87 & 7.44 & 6.51 & - & 8.2 & 15.5 & 23.0 & 4.7 & 13.3 & 62.41 \\
GPT-Image-1~\citep{gptimage} & 7.74 & 8.13 & 7.49 & 4.20 & 34.1 & 32.2 & 37.0 & 10.6 & 28.9 & 80.09 \\
GPT-Image-1.5~\citep{gptimage1.5} & - & - & - & 4.66 & 54.1 & 60.0 & 62.0 & 21.2 & 50.0 & - \\
Seedream-4.0~\citep{seedream4} & 8.14 & 8.12 & 7.70 & 4.30 & 12.9 & 12.2 & 11.0 & 7.1 & 10.8 & 77.31 \\
Seedream-4.5~\citep{seedream45} & 8.27 & 8.17 & 7.82 & 4.32 & - & - & - & - & - & - \\
Gemini 2.5 Flash Image~\citep{nanobanana} & 7.40 & 8.45 & 7.29 & 4.29 & 25.9 & 47.8 & 37.0 & 18.8 & 32.8 & 78.17 \\
Nano Banana Pro~\citep{nanobananapro} & 8.10 & 8.34 & 7.74 & 4.37 & 41.2 & 61.1 & 48.0 & 37.6 & 47.2 & 85.31 \\
\midrule
\rowcolor[HTML]{F5FFFA}
\multicolumn{11}{c}{\textit{EditThinker}} \\
OmniGen2$^{\ddag}$ & 6.47 & 7.04 & 6.03 & 3.41 & 2.4 & 1.1 & \textbf{7.0} & 1.2 & 3.1 & 50.52 \\
\hspace{1em} + EditThinker-8B & \textbf{6.65} & \textbf{7.44} & \textbf{6.42} & \textbf{3.67} & \textbf{3.5} & \textbf{10.0} & 5.0 & \textbf{3.5} & \textbf{5.6} & \textbf{53.09} \\
\hspace{1em} $\Delta$ & \etgain{0.18} & \etgain{0.40} & \etgain{0.39} & \etgain{0.26} & \etgain{1.1} & \etgain{8.9} & \etloss{2.0} & \etgain{2.3} & \etgain{2.5} & \etgain{2.57} \\
\midrule
FLUX.1-Kontext [Dev]$^{\ddag}$ & 6.62 & 7.61 & 6.18 & 3.36 & 2.3 & 5.5 & 13.0 & 1.2 & 5.8 & 61.81 \\
\hspace{1em} + EditThinker-8B & \textbf{7.59} & \textbf{7.63} & \textbf{7.02} & \textbf{4.07} & \textbf{12.9} & \textbf{16.6} & \textbf{25.0} & \textbf{5.8} & \textbf{15.5} & \textbf{69.53} \\
\hspace{1em} $\Delta$ & \etgain{0.97} & \etgain{0.02} & \etgain{0.84} & \etgain{0.71} & \etgain{10.6} & \etgain{11.1} & \etgain{12.0} & \etgain{4.6} & \etgain{9.7} & \etgain{7.72} \\
\midrule
Qwen-Image-Edit$^{\ddag}$ & 8.01 & \textbf{7.87} & 7.49 & 4.36 & 4.7 & 10.0 & 17.0 & 2.4 & 8.9 & 64.43 \\
\hspace{1em} + EditThinker-8B & \textbf{8.30} & 7.86 & \textbf{7.73} & \textbf{4.41} & \textbf{10.8} & \textbf{23.3} & \textbf{27.0} & \textbf{8.2} & \textbf{17.8} & \textbf{71.91} \\
\hspace{1em} $\Delta$ & \etgain{0.29} & \etloss{0.01} & \etgain{0.24} & \etgain{0.05} & \etgain{6.1} & \etgain{13.3} & \etgain{10.0} & \etgain{5.8} & \etgain{8.9} & \etgain{7.48} \\
\midrule
Longcat-Image-Edit$^{\ddag}$ & 8.25 & \textbf{8.17} & 7.81 & 4.53 & 9.4 & 15.6 & 17.0 & 4.7 & 11.9 & 67.52 \\
\hspace{1em} + EditThinker-8B & \textbf{8.44} & \textbf{8.17} & \textbf{8.00} & \textbf{4.57} & \textbf{27.1} & \textbf{28.9} & \textbf{24.0} & \textbf{8.2} & \textbf{22.2} & \textbf{75.00} \\
\hspace{1em} $\Delta$ & \etgain{0.19} & 0.00 & \etgain{0.19} & \etgain{0.04} & \etgain{17.7} & \etgain{13.3} & \etgain{7.0} & \etgain{3.5} & \etgain{10.3} & \etgain{7.48} \\
\bottomrule
\end{tabular}%
}
\end{minipage}
\par\vspace{\dbltextfloatsep}}]
\subsection{Main results}

\noindent\textbf{Performance on General Editing.}
Adding EditThinker-8B raises all four editors' overall scores on both GEdit-Bench-EN and ImgEdit-Bench (\cref{tab:Img-Bench_all}). FLUX.1-Kontext [Dev] gains the most on both benchmarks, rising from 6.18 to 7.02 on GEdit and from 3.36 to 4.07 on ImgEdit. ImgEdit gains are smaller for the two stronger single-pass editors, from 4.36 to 4.41 with Qwen-Image-Edit and from 4.53 to 4.57 with Longcat-Image-Edit.

On GEdit, semantic consistency ($G_{\mathrm{SC}}$) improves for all four editors. Perceptual quality ($G_{\mathrm{PQ}}$) improves most for OmniGen2, from 7.04 to 7.44. It rises from 7.61 to 7.63 for FLUX.1-Kontext and stays at 8.17 for Longcat-Image-Edit.

\noindent\textbf{Performance on Reasoning-Based Editing.}
On RISE-Bench, EditThinker-8B raises overall success rates by 2.5 to 10.3 percentage points across the four editors. Longcat-Image-Edit gains the most, from 11.9\% to 22.2\% overall, with temporal edits rising from 9.4\% to 27.1\%. With OmniGen2, causal edits improve from 1.1\% to 10.0\%. On KRIS-Bench, overall scores rise from 50.52 to 53.09 for OmniGen2, 61.81 to 69.53 for FLUX.1-Kontext [Dev], 64.43 to 71.91 for Qwen-Image-Edit, and 67.52 to 75.00 for Longcat-Image-Edit (\cref{tab:Img-Bench_all}).

\noindent\textbf{Comparison with ReasonEdit.}
In a separate comparison using Qwen-Image-Edit-2509, EditThinker-8B scores 7.85 on GEdit-Bench-EN and 4.58 on ImgEdit-Bench, compared with the published ReasonEdit scores of 7.77 and 4.36. ReasonEdit adapts the editor, while EditThinker uses the same fixed checkpoint as the single-pass baseline. The comparison uses the respective inference settings. The full comparison is in Appendix~\ref*{sec:additional_reasonedit}.

The reproduced single-pass baseline scores 7.51 on GEdit and 4.24 on ImgEdit. EditThinker-8B improves both scores by 0.34 with the same Qwen-Image-Edit-2509 checkpoint. The original Qwen-Image-Edit also improves on both benchmarks in \cref{tab:Img-Bench_all}, showing consistent gains from iterative refinement across the two releases.

\begin{figure*}[!t]
    \centering
    \includegraphics[pagebox=cropbox,width=0.97\linewidth]{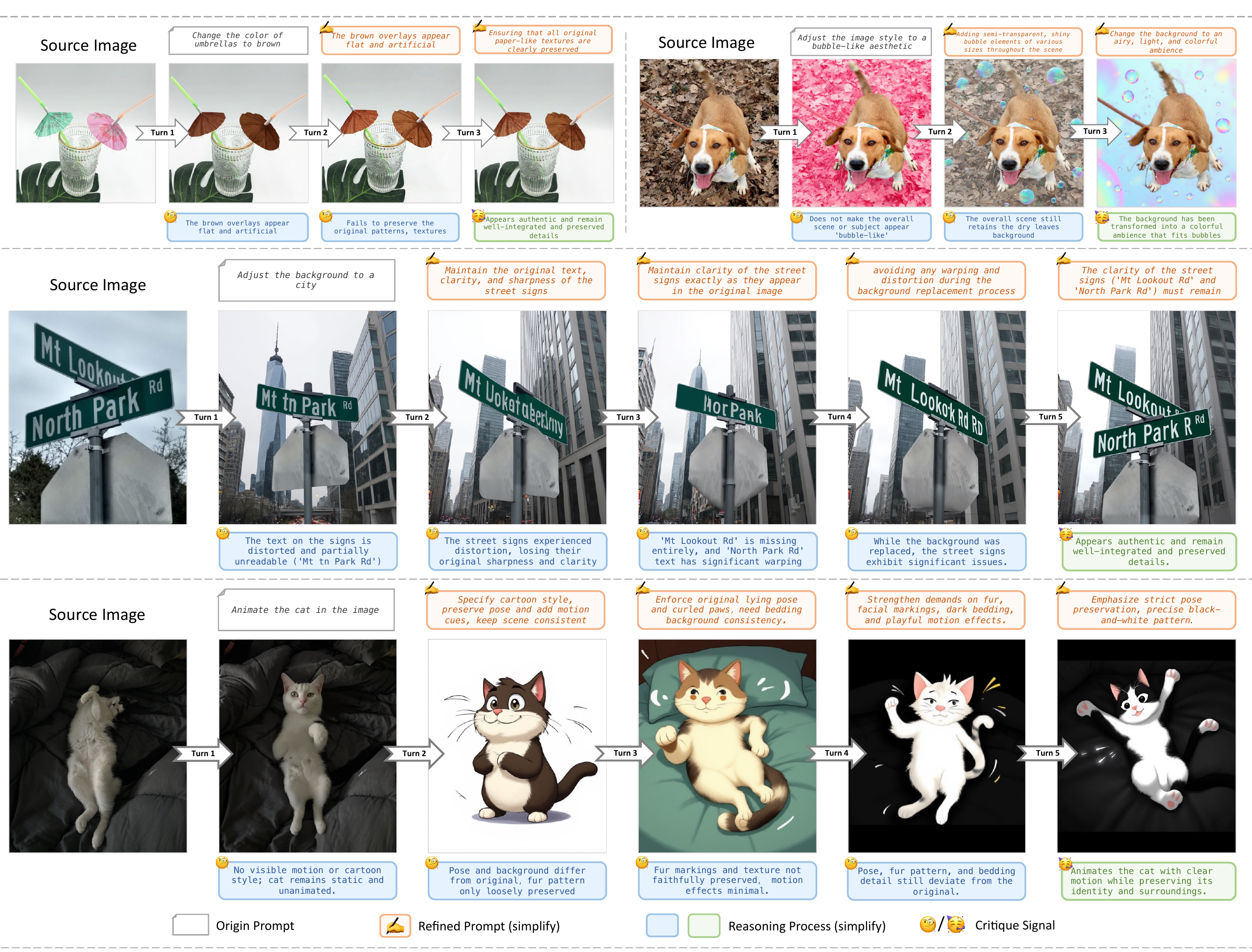}
    \caption{\textbf{Qualitative editing results.} Four trajectories with source images, edits, and abbreviated prompts and critiques.}
    \label{fig:qualitative_edits}
\end{figure*}

\subsection{Ablation study}

We use FLUX.1-Kontext [Dev] as the editor to study training, inference, thinker design, and backbone choices.

\noindent\textbf{Training Strategy.}
We test how SFT and the two RL rewards affect quality and editing rounds in \cref{tab:abl-sft-rl-Time-sub}. SFT improves GEdit from 6.43 to 6.92. Critic-only RL reduces rounds to 2.13 but lowers the score to 6.74; edit-only RL reaches 6.95 at 3.93 rounds. Combining both rewards reaches 7.02 at 2.19 rounds, improving both measures over SFT alone (6.92 at 2.55). The trade-off depends on the benchmark: edit-only RL scores higher on ImgEdit (4.15 versus 4.07), but requires more calls (3.70 versus 1.85). The GEdit score of 6.51 without SFT supports the cold start. Appendix~\ref*{sec:additional_reward_weights} reports the reward-weight ablations.

\begin{nolinenumbers}
\noindent\begin{minipage}{\columnwidth}
\centering
\captionsetup{hypcap=false}
\captionof{table}{SFT and RL rewards. Rounds are mean editor calls.}
\label{tab:abl-sft-rl-Time-sub}
\footnotesize
\setlength{\tabcolsep}{3pt}
\renewcommand{\arraystretch}{1.05}
\begin{adjustbox}{max width=\linewidth}
\begin{tabular}{c cc cccc cc}
\toprule
\multirow{2}{*}{SFT}
& \multicolumn{2}{c}{RL}
& \multicolumn{4}{c}{GEdit-Bench-EN}
& \multicolumn{2}{c}{ImgEdit-Bench} \\
\cmidrule(lr){2-3}\cmidrule(lr){4-7}\cmidrule(lr){8-9}
& $R_{critic}$ & $R_{edit}$
& G\_SC$\uparrow$ & G\_PQ$\uparrow$ & G\_O & Rounds$\downarrow$
& Overall$\uparrow$ & Rounds$\downarrow$ \\
\midrule
\multicolumn{3}{c}{FLUX.1-Kontext [Dev]} & 6.62 & 7.61 & 6.18 & 1.00 & 3.36 & 1.00 \\ \midrule
\xmark & \xmark & \xmark & 6.91 & 7.64 & 6.43 & 1.99 & 3.64 & 1.91 \\
\cmark & \xmark & \xmark & 7.53 & 7.58 & 6.92 & 2.55 & 3.98 & 2.24 \\
\cmark & \cmark & \xmark & 7.34 & 7.50 & 6.74 & 2.13 & 4.01 & 1.94 \\
\cmark & \xmark & \cmark & 7.43 & 7.65 & 6.95 & 3.93 & 4.15 & 3.70 \\
\xmark & \cmark & \cmark & 6.64 & 7.95 & 6.51 & 2.00 & 3.75 & 1.99 \\
\cmark & \cmark & \cmark & 7.59 & 7.63 & 7.02 & 2.19 & 4.07 & 1.85 \\
\bottomrule
\end{tabular}
\end{adjustbox}
\end{minipage}\par
\end{nolinenumbers}

\smallskip
\begin{nolinenumbers}
\noindent\begin{minipage}{\columnwidth}
\centering
\captionsetup{hypcap=false}
\captionof{table}{Editing-round budgets and average editor calls.}
\label{tab:abl-turn-models}
\footnotesize
\setlength{\tabcolsep}{3pt}
\renewcommand{\arraystretch}{1.05}
\begin{tabular*}{\linewidth}{@{\extracolsep{\fill}\hspace{\tabcolsep}}lcccc@{\hspace{\tabcolsep}}}
\toprule
\multirow{2}{*}{Max Turn}
& \multicolumn{2}{c}{EditThinker-8B}
& \multicolumn{2}{c}{EditScore-8B} \\
\cmidrule(lr){2-3}\cmidrule(lr){4-5}
& G\_O$\uparrow$ & Rounds$\downarrow$
& G\_O$\uparrow$ & Rounds$\downarrow$ \\
\midrule
1 & 6.18 & 1.00 & 6.18 & 1.00   \\
2 & 6.93 & 1.67 & 6.41 & 2.00   \\
4 & 7.05 & 2.07 & 6.59 & 4.00   \\
6 & 7.06 & 2.24 & 6.62 & 6.00   \\
8 & 7.14 & 2.35 & 6.64 & 8.00   \\
10 & 7.16 & 2.44 & 6.67 & 10.00   \\
\bottomrule
\end{tabular*}
\end{minipage}\par
\end{nolinenumbers}
\medskip

\noindent\textbf{Editing-Round Budget.}
\Cref{tab:abl-turn-models} compares iterative refinement with EditScore~\citep{luo2025editscore} best-of-$N$ sampling. EditThinker-8B gains 0.12 GEdit points from budgets 2 to 4, but only 0.02 from 8 to 10, indicating uneven marginal returns from additional budget. At budget 10, it scores 7.16 with 2.44 average calls, versus 6.67 with 10 calls for EditScore. Refinement thus outperforms selection while using fewer editor calls.

\par\smallskip
\begin{nolinenumbers}
\noindent\begin{minipage}{\columnwidth}
\centering
\captionsetup{hypcap=false}
\captionof{table}{Thinker design and data partition. Each SFT + RL run starts from its split's SFT checkpoint.}
\label{tab:additional_design_choices}
\footnotesize
\setlength{\tabcolsep}{3pt}
\renewcommand{\arraystretch}{1.05}
\begin{adjustbox}{max width=\linewidth}
\begin{tabular}{@{}lcccccc@{}}
\toprule
\multirow{2}{*}{Configuration} & \multicolumn{4}{c}{GEdit-EN} & \multicolumn{2}{c}{ImgEdit} \\
\cmidrule(lr){2-5}\cmidrule(lr){6-7}
& $G_{\mathrm{SC}}\uparrow$ & $G_{\mathrm{PQ}}\uparrow$ & $G_{\mathrm{O}}\uparrow$ & Rounds$\downarrow$ & Overall$\uparrow$ & Rounds$\downarrow$ \\
\midrule
Ours & 7.59 & 7.63 & 7.02 & 2.19 & 4.07 & 1.85 \\
\midrule
Decoupled roles & 7.42 & 7.61 & 6.90 & 2.45 & 3.88 & 2.03 \\
$T$-before-$R$ & 7.36 & 7.55 & 6.84 & 2.25 & 3.82 & 1.95 \\
w/o reasoning & 7.46 & 7.61 & 6.91 & 2.30 & 3.90 & 1.94 \\
\midrule
Random: SFT & 7.48 & 7.59 & 6.92 & 2.60 & 3.98 & 2.26 \\
Random: SFT + RL & 7.52 & 7.60 & 6.97 & 2.24 & 4.01 & 2.03 \\
Reversed: SFT & 7.49 & 7.60 & 6.95 & 2.50 & 3.95 & 2.19 \\
Reversed: SFT + RL & 7.38 & 7.56 & 6.87 & 2.36 & 3.84 & 2.08 \\
\bottomrule
\end{tabular}
\end{adjustbox}
\end{minipage}\par
\end{nolinenumbers}
\smallskip

\noindent\textbf{Thinker Design and Data Partition.}
We compare role coupling, output order, and explicit reasoning in \cref{tab:additional_design_choices}. Decoupling the Critic and Rewriter lowers GEdit/ImgEdit from 7.02/4.07 to 6.90/3.88 and increases average rounds. Generating the instruction before reasoning ($T$-before-$R$) yields 6.84/3.82. The variant without an explicit reasoning trace scores 6.91/3.90, with average rounds rising from 2.19/1.85 to 2.30/1.94.

The split ablations examine the assignment of stable or high-quality examples to SFT and high-variance \mbox{trajectories} to RL (\cref{sec:data_construction}). Random splitting improves during RL, from 6.92/3.98 to 6.97/4.01 on GEdit/ImgEdit. Reversing the allocation instead lowers scores from 6.95/3.95 to 6.87/3.84, even as average rounds decrease. The GEdit ranking of random and reversed splits therefore reverses after RL: the stronger SFT result does not imply a stronger final model. Both final variants trail our 7.02/4.07, favoring the proposed partition after RL.

\noindent\textbf{Generalization Across Training Backbones.}
\Cref{tab:additional_vlm_backbones} tests whether the training gains extend beyond the default backbone. Each trained variant is compared with its corresponding backbone without EditThinker training. All three improve on GEdit, ImgEdit, and RISE, with Qwen2.5-VL-7B rising from 6.26/3.44/10.0 to 6.93/4.06/12.7. The consistent within-backbone gains support applying the training procedure to different VLMs.

\begin{nolinenumbers}
\noindent\begin{minipage}{\columnwidth}
\centering
\captionsetup{hypcap=false}
\captionof{table}{Generalization across VLM backbones. ET denotes \mbox{EditThinker} training (\cmark: with; \xmark: without).}
\label{tab:additional_vlm_backbones}
\footnotesize
\setlength{\tabcolsep}{3pt}
\renewcommand{\arraystretch}{1.05}
\begin{tabular*}{\linewidth}{@{\extracolsep{\fill}\hspace{\tabcolsep}}lcccc@{\hspace{\tabcolsep}}}
\toprule
Backbone & ET & GEdit-EN & ImgEdit & RISE \\
\midrule
\multicolumn{2}{l}{FLUX.1-Kontext [Dev]} & 6.18 & 3.36 & 5.8 \\
\midrule
\multirow{2}{*}{InternVL2.5-8B} & \xmark & 6.22 & 3.38 & 8.1 \\
& \cmark & 6.84 & 3.54 & 11.1 \\
\midrule
\multirow{2}{*}{InternVL3-8B} & \xmark & 6.32 & 3.40 & 9.4 \\
& \cmark & 6.88 & 3.68 & 12.5 \\
\midrule
\multirow{2}{*}{Qwen2.5-VL-7B} & \xmark & 6.26 & 3.44 & 10.0 \\
& \cmark & 6.93 & 4.06 & 12.7 \\
\bottomrule
\end{tabular*}
\end{minipage}\par
\end{nolinenumbers}

\smallskip

\noindent\textbf{Evaluation with Gemini-2.5-Pro.}
We assess whether the training improvements also appear under a non-GPT judge in \cref{tab:additional_gemini_evaluation}. Gemini-2.5-Pro scores increase from 5.08/3.07 for the base FLUX editor to 5.38/3.23 with Qwen3-VL-8B without EditThinker training, 5.59/3.36 after SFT, and 5.74/3.44 after SFT and RL on GEdit/ImgEdit. The two judges rank these training stages in the same order, supporting the measured gains.

\par\medskip
\begin{nolinenumbers}
\noindent\begin{minipage}{\columnwidth}
\captionsetup{hypcap=false}
\centering
\captionof{table}{Gemini-2.5-Pro evaluation with FLUX.1-Kontext. Higher scores are better.}
\label{tab:additional_gemini_evaluation}
\small
\setlength{\tabcolsep}{3pt}
\begin{tabular*}{\linewidth}{@{\extracolsep{\fill}\hspace{\tabcolsep}}lcc@{\hspace{\tabcolsep}}}
\toprule
Configuration & GEdit-Bench-EN & ImgEdit-Bench \\
\midrule
FLUX.1-Kontext & 5.08 & 3.07 \\
+ Qwen3-VL-8B & 5.38 & 3.23 \\
+ SFT & 5.59 & 3.36 \\
+ SFT + RL & 5.74 & 3.44 \\
\bottomrule
\end{tabular*}
\end{minipage}\par
\end{nolinenumbers}
\medskip

\subsection{Comparison with prompt enhancement}
\Cref{tab:abl-think-paradigm} compares single-pass instruction enhancement with iterative refinement using the same fixed FLUX.1-Kontext editor. EditThinker-8B reaches a GEdit-Bench-EN overall score of 7.02, compared with 6.53 for PromptEnhancer-32B and 6.82 for GPT-4.1 prompt enhancement.

\par\medskip
\begin{nolinenumbers}
\noindent\begin{minipage}{\columnwidth}
\centering
\captionsetup{hypcap=false}
\captionof{table}{Prompt enhancement and iterative refinement on GEdit-Bench-EN. All methods use a fixed FLUX.1-Kontext [Dev] editor.}
\label{tab:abl-think-paradigm}
\small
\setlength{\tabcolsep}{3pt}
\begin{tabular*}{\linewidth}{@{\extracolsep{\fill}\hspace{\tabcolsep}}lccc@{\hspace{\tabcolsep}}}
\toprule
Method & $G_{\mathrm{SC}}\uparrow$ & $G_{\mathrm{PQ}}\uparrow$ & $G_{\mathrm{O}}\uparrow$ \\
\midrule
FLUX.1-Kontext [Dev] &6.62 & 7.61 & 6.18\\
\midrule
+ PromptEnhancer-32B & 6.94 & 7.62 & 6.53 \\
+ GPT-4.1 prompt enhancement &7.34 & 7.64 & 6.82 \\
+ EditThinker-8B refinement & 7.59 & 7.63 & 7.02 \\
\bottomrule
\end{tabular*}
\end{minipage}\par
\end{nolinenumbers}

\subsection{Qualitative analysis}
\Cref{fig:qualitative_edits} shows how critiques identify details to preserve. Naming the streets in a revised prompt restores legible signs after a background change. The umbrella edit retains the \mbox{paper texture} while changing color. Appendix~\ref*{E} shows more trajectories. Appendix~\ref*{app:qualitative} analyzes hard cases.

\section{Conclusion}
We present EditThinker, a plug-and-play 8B VLM that enhances fixed image editors through iterative reasoning. SFT and RL train the model to jointly critique results and refine instructions using expert demonstrations and executed editing feedback. EditThinker improves overall scores for all four editors on GEdit-Bench-EN, ImgEdit-Bench, RISE-Bench, and KRIS-Bench. In the round-budget study, it achieves higher GEdit scores with fewer editor calls than best-of-$N$ selection. Ablations support joint critique and refinement, reasoning before instruction generation, and the proposed allocation of data between SFT and RL. Consistent improvements across separately trained VLM backbones further suggest that the training approach extends beyond a single backbone. The gains are not uniform across subtasks, and some edits remain unresolved after further rounds.

\section*{LLM Usage Statement}
We train EditThinker from Qwen3-VL-8B-Instruct and compare it with other VLMs. \mbox{GPT-4.1} generates expert demonstrations for \textsc{ThinkEdit}-167k and scores edits for reinforcement learning (\cref{sec:data_construction,sec:rl}). In our experiments, \mbox{GPT-4.1} also provides single-pass prompt enhancement (\cref{tab:abl-think-paradigm}). Gemini-2.5-Pro provides an alternative evaluation of editing quality (\cref{tab:additional_gemini_evaluation}). We also used \mbox{OpenAI Codex} to write analysis and plotting code and to draft and edit parts of the text. The authors verified all code, results, and text and take full responsibility for the content of the paper.

{\small
\bibliographystyle{ieeenat_fullname}
\bibliography{references}
}

\clearpage
\onecolumn
\appendix
\setcounter{figure}{3}
\setcounter{table}{7}
\setcounter{equation}{5}
\addtocontents{toc}{\protect\setcounter{tocdepth}{2}}
\begin{center}
    \Large\bfseries Supplementary Material
\end{center}
\renewcommand{\contentsname}{Appendix Contents}
\pdfbookmark[0]{Appendix Contents}{appendix-contents}
\tableofcontents
\clearpage
\section{Additional Evaluation on KRIS-Bench}
\label{app:kris}

\Cref{tab:kris} gives the factual, conceptual, and procedural scores on KRIS-Bench. For each editor in the EditThinker group of Table~\ref*{tab:Img-Bench_all}, we evaluated the single-pass baseline and its EditThinker-8B counterpart on the same single-image subset. The baseline rows marked $^{\ddag}$ use our measurements with official checkpoints.

\definecolor{red4!20}{HTML}{F2B7B7} %
\definecolor{blue4!20}{HTML}{B7D7F2} %

\begin{table}[htbp]

\caption{Detailed KRIS-Bench results. The reference rows report official full-benchmark scores. $^{\ddag}$ denotes reproduced baselines. Temporal Prediction is recorded as $0.00^{*}$ in the paired evaluation because some editors lack multi-image input support.}
\centering
\label{tab:kris}
\newcommand{\kriszero}{0.00\rlap{\textsuperscript{*}}}

\newsavebox{\krisnumberbox}
\newlength{\krisnumberwidth}
\settowidth{\krisnumberwidth}{99.99}
\newcolumntype{K}{>{\begin{lrbox}{\krisnumberbox}}c<{\end{lrbox}\makebox[\krisnumberwidth][r]{\usebox{\krisnumberbox}}}}
\setlength{\tabcolsep}{3pt}
\resizebox{\linewidth}{!}{%
\begin{tabular}{@{}l*{10}{K}|K@{}}
\toprule
\multirow{2}{*}{\textbf{Model}} & \multicolumn{4}{c}{\textbf{Factual Knowledge}} & \multicolumn{3}{c}{\textbf{Conceptual Knowledge}} & \multicolumn{3}{c}{\textbf{Procedural Knowledge}} & \multicolumn{1}{|c@{}}{\multirow{2}{*}{\textbf{Overall}}} \\
\cmidrule(lr){2-5} \cmidrule(lr){6-8} \cmidrule(lr){9-11}
& \multicolumn{1}{c}{\textbf{Attribute}} & \multicolumn{1}{c}{\textbf{Spatial}} & \multicolumn{1}{c}{\textbf{Temporal}} & \multicolumn{1}{c}{\textbf{Average}} & \multicolumn{1}{c}{\textbf{Social Sci.}} & \multicolumn{1}{c}{\textbf{Natural Sci.}} & \multicolumn{1}{c}{\textbf{Average}} & \multicolumn{1}{c}{\textbf{Logical}} & \multicolumn{1}{c}{\textbf{Instruction}} & \multicolumn{1}{c}{\textbf{Average}} & \\
\midrule \rowcolor{red4!20}

\multicolumn{12}{c}{\textit{Proprietary Models}} \\
Doubao & 70.92 & 59.17 & 40.58 & 63.30 & 65.50 & 61.19 & 62.23 & 47.75 & 60.58 & 54.17 & 60.70 \\
Step 3o vision & 69.67 & 61.08 & 63.25 & 66.70 & 66.88 & 60.88 & 62.32 & 49.06 & 54.92 & 51.99 & 61.43 \\
Gemini 2.0 & 66.33 & 63.33 & 63.92 & 65.26 & 68.19 & 56.94 & 59.65 & 54.13 & 71.67 & 62.90 & 62.41 \\
GPT-Image-1 & 83.17 & 79.08 & 68.25 & 79.80 & 85.50 & 80.06 & 81.37 & 71.56 & 85.08 & 78.32 & 80.09 \\
\midrule \rowcolor{blue4!20}

\multicolumn{12}{c}{\textit{Open-source Models}} \\
InstructPix2Pix & 30.33 & 21.33 & 0.00 & 23.33 & 22.56 & 26.56 & 25.59 & 19.81 & 14.75 & 17.28 & 22.82 \\
OmniGen & 37.92 & 28.25 & 21.83 & 33.11 & 30.63 & 27.19 & 28.02 & 11.94 & 35.83 & 23.89 & 28.85 \\
MagicBrush & 53.92 & 39.58 & 0.00 & 41.84 & 42.94 & 38.06 & 39.24 & 30.00 & 23.08 & 26.54 & 37.15 \\
AnyEdit & 47.67 & 45.17 & 0.00 & 39.26 & 38.56 & 42.94 & 41.88 & 36.56 & 26.92 & 31.74 & 38.55 \\
Emu2 & 51.50 & 45.83 & 22.17 & 45.40 & 34.69 & 38.44 & 37.54 & 24.81 & 45.00 & 34.91 & 39.70 \\
Step1X-Edit & 55.50 & 51.75 & 0.00 & 45.52 & 44.69 & 49.06 & 48.01 & 40.88 & 22.75 & 31.82 & 43.29 \\
HiDream-E1 & 52.75 & 49.42 & 0.00 & 43.31 & 52.56 & 49.25 & 50.05 & 45.19 & 30.08 & 37.64 & 44.72 \\
ByteMorph & 61.17 & 62.00 & 0.00 & 51.27 & 45.50 & 47.38 & 46.92 & 32.00 & 31.33 & 31.67 & 44.85 \\
FLUX.1-Kontext [Dev] & 64.83 & 60.92 & 0.00 & 53.28 & 48.94 & 50.81 & 50.36 & 46.06 & 39.00 & 42.53 & 49.54 \\
OmniGen2 & 59.92 & 52.25 & 54.75 & 57.36 & 47.56 & 43.12 & 44.20 & 32.50 & 63.08 & 47.79 & 49.71 \\
UniWorld-V1 & 58.17 & 54.50 & 63.00 & 58.15 & 47.50 & 43.94 & 44.80 & 42.00 & 53.83 & 47.92 & 50.27 \\
Step1X-Edit v1.1 & 64.17 & 61.75 & 0.00 & 53.05 & 52.06 & 55.06 & 54.34 & 52.56 & 36.75 & 44.66 & 51.59 \\
BAGEL & 64.27 & 62.42 & 42.45 & 60.26 & 55.40 & 56.01 & 55.86 & 52.54 & 50.56 & 51.55 & 56.40 \\
BAGEL-Think  & 67.42 & 68.33 & 58.67 & 66.18 & 63.55 & 61.40 & 61.92 & 48.12 & 50.22 & 49.17 & 60.41 \\

\midrule \rowcolor[HTML]{F5FFFA}
\multicolumn{12}{c}{\textit{EditThinker}} \\
OmniGen2$^{\ddag}$ & 60.21 & 54.67 & \kriszero & 58.73 & 53.60 & 46.76 & 48.42 & 37.67 & 56.67 & 45.84 & 50.52 \\
\hspace{1em} + EditThinker-8B & 62.18 & 53.92 & \kriszero & 59.98 & 61.50 & 49.90 & 52.71 & 37.04 & 58.78 & 46.43 & 53.09 \\
\midrule
FLUX.1-Kontext [Dev]$^{\ddag}$ & 71.12 & 67.25 & \kriszero & 70.09 & 56.60 & 58.91 & 58.35 & 56.75 & 63.72 & 59.75 & 61.81 \\
\hspace{1em} + EditThinker-8B & 77.82 & 65.50 & \kriszero & 73.73 & 73.44 & 69.04 & 70.09 & 62.29 & 65.33 & 63.60 & 69.53 \\
\midrule
Qwen-Image-Edit$^{\ddag}$ & 72.73 & 73.33 & \kriszero & 72.89 & 63.50 & 60.40 & 61.15 & 57.47 & 67.97 & 61.70 & 64.43 \\
\hspace{1em} + EditThinker-8B & 78.48 & 73.83 & \kriszero & 77.24 & 76.20 & 70.69 & 72.02 & 65.23 & 66.89 & 65.94 & 71.91 \\
\midrule
Longcat-Image-Edit$^{\ddag}$ & 77.67 & 75.25 & \kriszero & 77.02 & 65.50 & 63.12 & 63.69 & 61.79 & 69.00 & 64.88 & 67.52 \\
\hspace{1em} + EditThinker-8B & 82.61 & 80.33 & \kriszero & 82.00 & 77.30 & 73.31 & 74.28 & 68.04 & 70.11 & 68.93 & 75.00 \\
\bottomrule
\end{tabular}%
}
\end{table}

\clearpage
\section{Additional Ablation Studies}
\label{app:additional_experiments}

This section reports the full ReasonEdit comparison and reward-weight ablations. The thinker-design, data-partition, VLM-backbone, alternative-judge, and prompt-enhancement experiments are presented in the main paper (\cref*{tab:additional_design_choices,tab:additional_vlm_backbones,tab:additional_gemini_evaluation,tab:abl-think-paradigm}). Where efficiency is reported, \emph{average rounds} denotes the mean number of editing iterations. It serves as a proxy for editor calls.

\subsection{Comparison with ReasonEdit}
\label{sec:additional_reasonedit}

\Cref{tab:additional_reasonedit} compares ReasonEdit and EditThinker-8B using Qwen-Image-Edit-2509 as the base editor. We evaluate the single-pass baseline and EditThinker-8B with the same fixed checkpoint. The ReasonEdit scores come from the published evaluation of its adapted editor under its own inference settings. On GEdit-Bench-EN and ImgEdit-Bench, the single-pass baseline scores 7.51 and 4.24, EditThinker-8B scores 7.85 and 4.58, and ReasonEdit scores 7.77 and 4.36.

\begin{table}[htbp]
\centering
\caption{Comparison on GEdit-Bench-EN and ImgEdit-Bench with Qwen-Image-Edit-2509 as the base editor. The baseline and EditThinker-8B results are from our evaluation; ReasonEdit results are taken from the published paper. Higher scores are better.}
\label{tab:additional_reasonedit}
\small
\setlength{\tabcolsep}{5pt}
\begin{tabular}{@{}lcc@{}}
\toprule
Method & GEdit-Bench-EN & ImgEdit-Bench \\
\midrule
Qwen-Image-Edit-2509 & 7.51 & 4.24 \\
\quad + ReasonEdit & 7.77 & 4.36 \\
\quad + EditThinker-8B & 7.85 & 4.58 \\
\bottomrule
\end{tabular}
\end{table}

\subsection{Reward-Weight Sensitivity}
\label{sec:additional_reward_weights}

\Cref{tab:additional_reward_weights} varies the weights of the format, critic, and edit rewards. The selected setting $(\alpha,\beta,\gamma)=(0.50,0.10,0.40)$ yields a GEdit-Bench-EN overall score of 7.02 at 2.19 average rounds and an ImgEdit-Bench score of 4.07 at 1.85 average rounds. Removing the critic weight while increasing the edit weight to 0.50 yields a higher ImgEdit-Bench score of 4.15, with 3.70 average rounds; its GEdit-Bench-EN score is 6.95 at 3.93 rounds. These measurements illustrate a quality-iteration trade-off among the tested settings.

The $(\alpha,\beta,\gamma)=(0.25,0.15,0.60)$ configuration yields lower overall scores on both benchmarks than the selected setting. In the $\alpha=0.50$ sweep, we vary the critic weight $\beta$ from 0 to 0.50 and set the edit weight to $\gamma=0.50-\beta$.

\begin{table}[htbp]
\centering
\caption{Reward-weight ablation. Weights $(\alpha,\beta,\gamma)$ apply to the format, critic, and edit rewards, respectively. Avg. rounds is the reported average number of editing iterations, used as an editor-call proxy.}
\label{tab:additional_reward_weights}
\small
\setlength{\tabcolsep}{3pt}
\begin{tabular}{@{}ccccccccc@{}}
\toprule
\multicolumn{3}{c}{Reward weights} & \multicolumn{4}{c}{GEdit-Bench-EN} & \multicolumn{2}{c}{ImgEdit-Bench} \\
\cmidrule(lr){1-3}\cmidrule(lr){4-7}\cmidrule(lr){8-9}
$\alpha$ & $\beta$ & $\gamma$ & $G_{\mathrm{SC}}\uparrow$ & $G_{\mathrm{PQ}}\uparrow$ & $G_{\mathrm{O}}\uparrow$ & \shortstack{Avg.\\rounds$\downarrow$} & Overall$\uparrow$ & \shortstack{Avg.\\rounds$\downarrow$} \\
\midrule
0.50 & 0.10 & 0.40 & 7.59 & 7.63 & 7.02 & 2.19 & 4.07 & 1.85 \\
\midrule
0.25 & 0.15 & 0.60 & 7.22 & 7.59 & 6.70 & 2.03 & 3.84 & 1.98 \\
0.50 & 0.00 & 0.50 & 7.43 & 7.65 & 6.95 & 3.93 & 4.15 & 3.70 \\
0.50 & 0.20 & 0.30 & 7.44 & 7.64 & 6.96 & 2.16 & 4.05 & 1.89 \\
0.50 & 0.30 & 0.20 & 7.40 & 7.55 & 6.88 & 2.13 & 4.06 & 1.94 \\
0.50 & 0.40 & 0.10 & 7.39 & 7.54 & 6.87 & 2.14 & 4.02 & 1.96 \\
0.50 & 0.50 & 0.00 & 7.34 & 7.50 & 6.74 & 2.13 & 4.01 & 1.94 \\
\bottomrule
\end{tabular}
\end{table}

\clearpage

\section{Additional Implementation Details}
\label{C}

\subsection{Details of EditThinker Expert}
To supervise EditThinker with high-quality reasoning traces and refined editing instructions, we employ GPT-4.1 as the \textit{expert} model. At the $t$-th editing iteration, we provide the tuple $(I_{\text{src}},\, I^{t-1}_{\text{edit}},\, T_s,\, T_{t-1})$ as input to the expert. The expert returns a JSON object containing \texttt{reason}, \texttt{new\_rewritten\_prompt}, and the Boolean field \texttt{is\_satisfied}. The first two fields provide $R_t$ and $T_t$ when refinement is needed. In the trajectory-generation notation, \texttt{<stop>} denotes \texttt{is\_satisfied = true}. This value ends the trajectory and sets \texttt{new\_rewritten\_prompt} to \texttt{null}. When \texttt{is\_satisfied = false}, the expert supplies the next instruction and the editor continues. The expert prompt is meticulously designed to explicitly encourage a multi-step cycle. It requires the model to: (1) evaluate whether the edited image fulfills the original instruction $T_s$, (2) identify failure causes through detailed reasoning, and (3) synthesize an improved instruction that corrects these errors without introducing new inconsistencies. The full prompt template used for the expert is provided in \cref{fig:expert_prompt}.

\subsection{Details of EditThinker}
EditThinker adopts a unified prompt format for both training and inference. This design ensures that the behavior learned during supervision aligns seamlessly with the capabilities required at inference time, enabling the model to (1) evaluate the current result, (2) reason about potential issues, and (3) refine the instruction for the next round. 

At each iteration $t$, EditThinker receives a multimodal tuple $(I_{\text{src}}, I_{\text{edit}}^{t-1}, T_s, T_{t-1})$ that provides the complete context of the editing state. Here, $I_{\text{src}}$ and $T_s$ represent the original source image and user instruction; $I_{\text{edit}}^{t-1}$ denotes the intermediate result from the previous turn; and $T_{t-1}$ is the specific instruction that produced it. Based on this input, EditThinker emits three tagged fields in the order \texttt{<think>}, \texttt{<score>}, and \texttt{<answer>}. They contain the reasoning trace $R_t$, the semantic and quality scores, and the refined instruction $T_t$, respectively. The main text abbreviates the scores as the ordered pair $[S_{\mathrm{sem},t}, S_{\mathrm{qual},t}]$. In the full prompt, \texttt{<score>} contains a JSON object with the keys \texttt{semantic} and \texttt{quality}, each holding a numeric value from 0 to 10. The inference controller reads these two fields and applies the stopping rule below. This offers two key advantages: (1) it provides a smoother, more informative supervision signal for learning nuanced failure patterns; and (2) it enables precise control over inference quality, allowing users to trigger refinement only when the predicted score falls below a specific threshold. The full prompt template used for the EditThinker is provided in \cref{fig:edit_thinker_prompt}.

\subsection{Hyperparameters and Stopping Criterion}
\label{sec:additional_stopping}
We use reward weights $(\alpha,\beta,\gamma)=(0.5,0.1,0.4)$ for format, critique, and edit rewards, respectively. The trajectory score-variance threshold is $\theta=3.0$. At inference, early stopping uses the geometric mean of the two predicted scores:
\begin{equation}
  S_{\mathrm{stop},t}=\sqrt{S_{\mathrm{sem},t}\,S_{\mathrm{qual},t}},
  \qquad \text{stop if }S_{\mathrm{stop},t}\geq 8.
\end{equation}
The default maximum is $N=5$ editing rounds. This inference budget is distinct from the GRPO rollout group size of 8 used during training. The benchmark results use this five-round default unless a round-budget ablation specifies another limit. The extended qualitative sequences in \cref{fig:additional_failure_cases}(c,d) are shown through 16 editing rounds to examine persistent errors. Reward-weight ablations appear in Appendix~\ref{sec:additional_reward_weights}, and design ablations in \cref*{tab:additional_design_choices}.

\clearpage
\setcounter{figure}{5}
\subsection{Prompt templates}
\label{sec:prompt_templates}
The expert uses the satisfaction Boolean described in Appendix~C.1. The trained Thinker uses the tagged response described in Appendix~C.2 and the score-based stopping rule in Appendix~C.3.

\begingroup
\begin{tcolorbox}[
    enhanced,
    breakable,
    colback=gray!10,
    colframe=black,
    boxrule=0.8pt,
    arc=2pt,
    left=5pt, right=5pt, top=6pt, bottom=6pt
]

\small
\raggedright

\noindent You are an expert image editing evaluator and prompt engineer. Your task is to:
\begin{enumerate}[leftmargin=14pt,itemsep=1pt,topsep=2pt]
    \item Evaluate whether an edited image successfully fulfills the original user instruction.
    \item If not satisfied, generate an improved rewritten prompt that addresses the shortcomings.
\end{enumerate}

\vspace{6pt}
\noindent\textbf{Input Information.}
You will receive:
\begin{itemize}[leftmargin=14pt,itemsep=1pt]
    \item \textbf{Original Image}: the input image before editing.
    \item \textbf{Original User Instruction}: the user's initial editing request.
    \item \textbf{Rewritten Prompt}: the refined instruction that was used for editing.
    \item \textbf{Edited Image}: the resulting image after applying the rewritten prompt.
\end{itemize}

\vspace{6pt}
\noindent\textbf{Evaluation Criteria.}

\noindent\textit{A. Intent Alignment}
\begin{itemize}[leftmargin=14pt,itemsep=1pt]
    \item Does the edited image achieve the core goal of the user instruction?
    \item Are all requested changes present and correctly implemented?
\end{itemize}

\noindent\textit{B. Quality Assessment}
\begin{itemize}[leftmargin=14pt,itemsep=1pt]
    \item \textbf{Subject/Object Changes}: correctness of additions/removals/replacements.
    \item \textbf{Appearance Modifications}: accuracy of color/style/material edits.
    \item \textbf{Scene Changes}: correctness of background/environment edits.
    \item \textbf{Detail Preservation}: important details remain intact.
    \item \textbf{Visual Coherence}: edited image looks natural and well-integrated.
\end{itemize}

\noindent\textit{C. Common Failure Patterns}
\begin{itemize}[leftmargin=14pt,itemsep=1pt]
    \item Missing requested elements.
    \item Incorrect positioning or scale.
    \item Wrong colors or materials.
    \item Unnatural blending or artifacts.
    \item Lost subject details.
    \item Style inconsistency.
    \item Text errors (if applicable).
    \item Over-editing or under-editing.
\end{itemize}

\vspace{6pt}
\noindent\textbf{Evaluation Decision.}

\noindent\textbf{SATISFIED}: the edited image fulfills the original instruction with acceptable quality.  
Minor imperfections are acceptable if the core intent is achieved.

\noindent\textbf{NOT SATISFIED}: the edit fails in key aspects.  
Major elements missing, incorrect, or severe quality issues $\rightarrow$ refinement required.

\vspace{6pt}
\noindent\textbf{Prompt Refinement Strategy (If Not Satisfied).}

\noindent\textit{1. Identify what went wrong.}
\begin{itemize}[leftmargin=14pt,itemsep=1pt]
    \item Compare original instruction $\rightarrow$ rewritten prompt $\rightarrow$ edited result.
    \item Identify mismatches between intent and execution.
    \item Determine whether the issue is clarity, specificity, or contradiction.
\end{itemize}

\noindent\textit{2. Refinement approaches.}
\begin{itemize}[leftmargin=14pt,itemsep=1pt]
    \item \textbf{If vague:} add specific descriptors, spatial relations, or context.
    \item \textbf{If contradictory:} resolve conflicts and simplify.
    \item \textbf{If important details were lost:} explicitly require preservation.
    \item \textbf{If scale/position wrong:} add precise location and size cues.
    \item \textbf{If style incorrect:} specify textures, lighting, materials.
    \item \textbf{If over/under-edited:} specify degree of modification.
\end{itemize}

\noindent\textit{3. Leverage all information.}
\begin{itemize}[leftmargin=14pt,itemsep=1pt]
    \item Reference visible content in the original and edited images.
    \item Retain what worked; correct what failed.
\end{itemize}

\vspace{6pt}
\noindent\textbf{Output Format.}\par\nopagebreak[4]

\begin{tcolorbox}[
    colback=white,
    colframe=gray!60,
    boxrule=0.5pt,
    left=4pt, right=4pt, top=3pt, bottom=3pt
]
\ttfamily\small
\{ \\
\ \ "is\_satisfied": true/false, \\
\ \ "reason": "Detailed explanation of evaluation. If satisfied, explain why it meets requirements. If not satisfied, describe specific shortcomings.",\\
\ \ "new\_rewritten\_prompt": "Use the refined instruction if is\_satisfied is false; otherwise, set to null."\\
\}
\end{tcolorbox}

\vspace{4pt}
\noindent\textbf{Examples.}

\noindent\textbf{Example 1: Satisfied}
\begin{tcolorbox}[colback=white,colframe=gray!60,boxrule=0.5pt,left=4pt,right=4pt]
\ttfamily\small
\{ "is\_satisfied": true, \\
\ \ "reason": "The edited image successfully adds a cat...", \\
\ \ "new\_rewritten\_prompt": null \}
\end{tcolorbox}

\noindent\textbf{Example 2: Not Satisfied — Lack of Specificity}
\begin{tcolorbox}[colback=white,colframe=gray!60,boxrule=0.5pt,left=4pt,right=4pt]
\ttfamily\small
\{ "is\_satisfied": false, \\
\ \ "reason": "The rewritten prompt was too vague...", \\
\ \ "new\_rewritten\_prompt": "Change the car color to blue..." \}
\end{tcolorbox}

\noindent\textbf{Input Data: }
\begin{itemize}[leftmargin=14pt,itemsep=2pt]
    \item \textbf{Original User Instruction}: \textcolor{blue}{$T_{s}$}
    \item \textbf{Rewritten Prompt Used}: \textcolor{blue}{$T_{t-1}$}
    \item \textbf{Images Order}: [Original Image, Edited Image]
\end{itemize}

Images: \textcolor{blue}{$I_{src}$}, \textcolor{blue}{$I_{edit}^{t-1}$}

\end{tcolorbox}

\captionsetup{hypcap=false}
\captionof{figure}{\textbf{EditThinker Expert Prompt.}  
The full expert instruction used for EditThinker Expert.  At each iteration, the Expert observes $(I_{src}, I_{edit}^{t-1}, T_s, T_{t-1})$ and returns reasoning, a satisfaction Boolean, and a refined instruction. A true satisfaction Boolean ends the trajectory and sets the new instruction to null. }
\label{fig:expert_prompt}
\endgroup

\clearpage
\begingroup
\begin{tcolorbox}[
    enhanced,
    breakable,
    colback=gray!10,
    colframe=black,
    boxrule=0.8pt,
    arc=2pt,
    left=4pt, right=4pt, top=5pt, bottom=5pt
]

\small
\raggedright

Images: \textcolor{blue}{$I_{src}$},  \textcolor{blue}{$I_{edit}^{t-1}$}
\vspace{6pt}

\noindent\textbf{Edit Evaluation and Prompt Refinement System.} 

You are an expert image editing evaluator and prompt engineer. Your task is to:
\begin{enumerate}[leftmargin=12pt,itemsep=1pt,topsep=2pt]
    \item Score the edited image from two perspectives and output a JSON object with numeric \texttt{semantic} and \texttt{quality} fields inside \texttt{<score>} tags.
    \item If you think the edited image is not good enough, generate an optimized rewritten prompt that addresses the original shortcomings; if you think it is good enough, output the original rewritten prompt.
\end{enumerate}

\vspace{6pt}
\noindent\textbf{Input Information.}
You are shown two images in sequence:
\begin{itemize}[leftmargin=12pt,itemsep=1pt,topsep=2pt]
    \item \textbf{Original Image}: the input image before editing.
    \item \textbf{Edited Image}: the latest edited image generated using the previous instruction.
\end{itemize}
The textual instructions involved in this process are:
\begin{itemize}[leftmargin=12pt,itemsep=1pt,topsep=2pt]
    \item \textbf{Original User Instruction} : ``\textcolor{blue}{$T_s$}"
    \item \textbf{Previous Rewritten Instruction} : ``\textcolor{blue}{$T_{t-1}$}"
\end{itemize}

\vspace{6pt}
\noindent\textbf{Evaluation Criteria (Score 0--10).}

\noindent\textit{A. Semantic Score (Instruction Following and Preservation).}
\begin{itemize}[leftmargin=12pt,itemsep=1pt,topsep=2pt]
    \item Evaluates how accurately the edit was performed. The edit fails if it either (A) fails to follow the text instruction, or (B) over-edits the image by changing content that was not supposed to be changed.
    \item $(10)$ Follows the instruction perfectly \emph{and} preserves all unchanged content.
    \item $(0)$ Fails to follow the instruction \emph{or} needlessly changes the original scene.
\end{itemize}

\noindent\textit{B. Quality Score (Naturalness \& Artifacts).}
\begin{itemize}[leftmargin=12pt,itemsep=1pt,topsep=2pt]
    \item Evaluates the technical quality of the newly edited image.
    \item $(10)$ Looks natural, has no artifacts, and integrates seamlessly.
    \item $(0)$ Looks unnatural (wrong shadow, lighting, or sense of distance) or contains severe artifacts (distortion, blurred faces, unusual body parts, disharmony).
\end{itemize}

\vspace{6pt}
\noindent\textbf{Prompt Refinement Strategy (if Not Good Enough).}
When generating a new rewritten prompt, follow these steps:

\noindent\textit{1. What went wrong?}
\begin{itemize}[leftmargin=14pt,itemsep=1pt,topsep=2pt]
    \item Compare original instruction $\rightarrow$ rewritten prompt $\rightarrow$ edited result.
    \item Identify gaps between intent and execution.
    \item Determine if the issue is clarity, specificity, or contradiction.
\end{itemize}

\noindent\textit{2. Refinement approaches.}
\begin{itemize}[leftmargin=14pt,itemsep=1pt,topsep=2pt]
    \item \textbf{If the rewritten prompt was too vague:}  
    add more specific descriptors (exact colors, positions, sizes), include spatial relationships and context, and specify interaction with existing elements.
    \item \textbf{If the rewritten prompt was contradictory:}  
    resolve conflicts between requirements, prioritize core intent over secondary details, and simplify complex multi-part instructions.
    \item \textbf{If important details were lost:}  
    explicitly state preservation requirements, add ``maintain [aspect]" or ``preserve [feature]" clauses, and reference specific elements from the original image.
    \item \textbf{If positioning/scale was wrong:}  
    use more precise spatial descriptors, add relative size/scale indicators, and specify foreground/midground/background placement.
    \item \textbf{If style/appearance was incorrect:}  
    use more specific visual vocabulary, add reference to the original image's style elements, and include material/texture/lighting specifications.
    \item \textbf{If the edit was over/under-processed:}  
    add modifiers such as ``subtle", ``gentle", ``dramatic", or ``significant", specify the degree of change more clearly, and balance enhancement with naturalness.
\end{itemize}

\noindent\textit{3. Leverage all information.}
\begin{itemize}[leftmargin=14pt,itemsep=1pt,topsep=2pt]
    \item Reference what is visible in the original image.
    \item Learn from what the previous rewritten prompt missed.
    \item Use the edited image as feedback on what went wrong.
    \item Maintain what worked, and only modify what needs to be improved.
\end{itemize}

\vspace{6pt}
\noindent\textbf{Output Format.}\par\nopagebreak[4]
The output consists of three parts:
\begin{enumerate}[leftmargin=12pt,itemsep=1pt,topsep=2pt]
    \item A statement: analysis process and reasoning.
    \item A JSON object: numeric \texttt{semantic} and \texttt{quality} scores, each ranging from 0 to 10.
    \item A prompt: either the optimized rewritten prompt or the original rewritten prompt.
\end{enumerate}

\vspace{2pt}
\noindent An illustrative response is shown below. The numeric scores illustrate the output format.
\begin{tcolorbox}[
    colback=white,
    colframe=gray!60,
    left=4pt, right=4pt, top=3pt, bottom=3pt,
    boxrule=0.5pt
]
\ttfamily\small
<think> \\
The car is only partly blue. Its shape and background are preserved. \\
</think>

<score> \\
\{\\
\ \ "semantic": 7.0,\\
\ \ "quality": 8.0\\
\}\\
</score>

<answer> \\
Change the car color to blue while preserving its shape, details, and background.\\
</answer>
\end{tcolorbox}

\end{tcolorbox}

\captionsetup{hypcap=false}
\captionof{figure}{\textbf{EditThinker prompt.} The unified prompt template used for EditThinker’s edit evaluation and instruction refinement. At each iteration, the Thinker observes $(I_{src}, I_{edit}^{t-1}, T_s, T_{t-1})$ and returns reasoning, a JSON score object with semantic and quality fields, and a refined instruction in that order.}
\label{fig:edit_thinker_prompt}
\endgroup

\clearpage

\section{ Details of \textsc{ThinkEdit-167K} Dataset}
\label{D}

We obtain $T_s$ and $I_{\text{src}}$ from three data sources: {OpenGPT-4o-Image}, {ShareGPT-4o-Image}, and {Pico-Banana-400K}. From these sources, we sample 40K, 40K, and 60K editing instances respectively, ensuring that the editing categories are as evenly distributed as possible, resulting in a total of 140K raw samples. We divide these samples into three splits and use {GPT-4.1} as the {EditThinker-Expert}, while selecting {OmniGen2}, {FLUX.1 Kontext [Dev]}, and {Qwen-Image-Edit} as the editors. After trajectory filtering, we retain 70K valid trajectories. Among them, 10K trajectories are selected for RL, while the remaining 60K undergo step-wise refinement and filtering, ultimately producing 140K high-quality samples for SFT. Unrolling the 10K selected RL trajectories yields 27K step-wise samples for RL training.

\section{Visualization}
\label{E}
We provide visualizations of the outputs generated by our framework across different settings. As shown in \cref{fig:vis-all}, EditThinker produces high-quality edits when combined with various editors, including FLUX.1 Kontext [Dev], OmniGen2, and Qwen-Image-Edit. In addition, \cref{fig:vis-reason} visualizes EditThinker’s iterative reasoning dynamics with FLUX.1 Kontext [Dev], highlighting how the Thinker critiques intermediate results and progressively refines the instruction over multiple rounds.

\setcounter{figure}{3}
\begin{figure}[p]
    \centering 
    \includegraphics[width=0.9\linewidth,height=0.89\textheight,keepaspectratio]{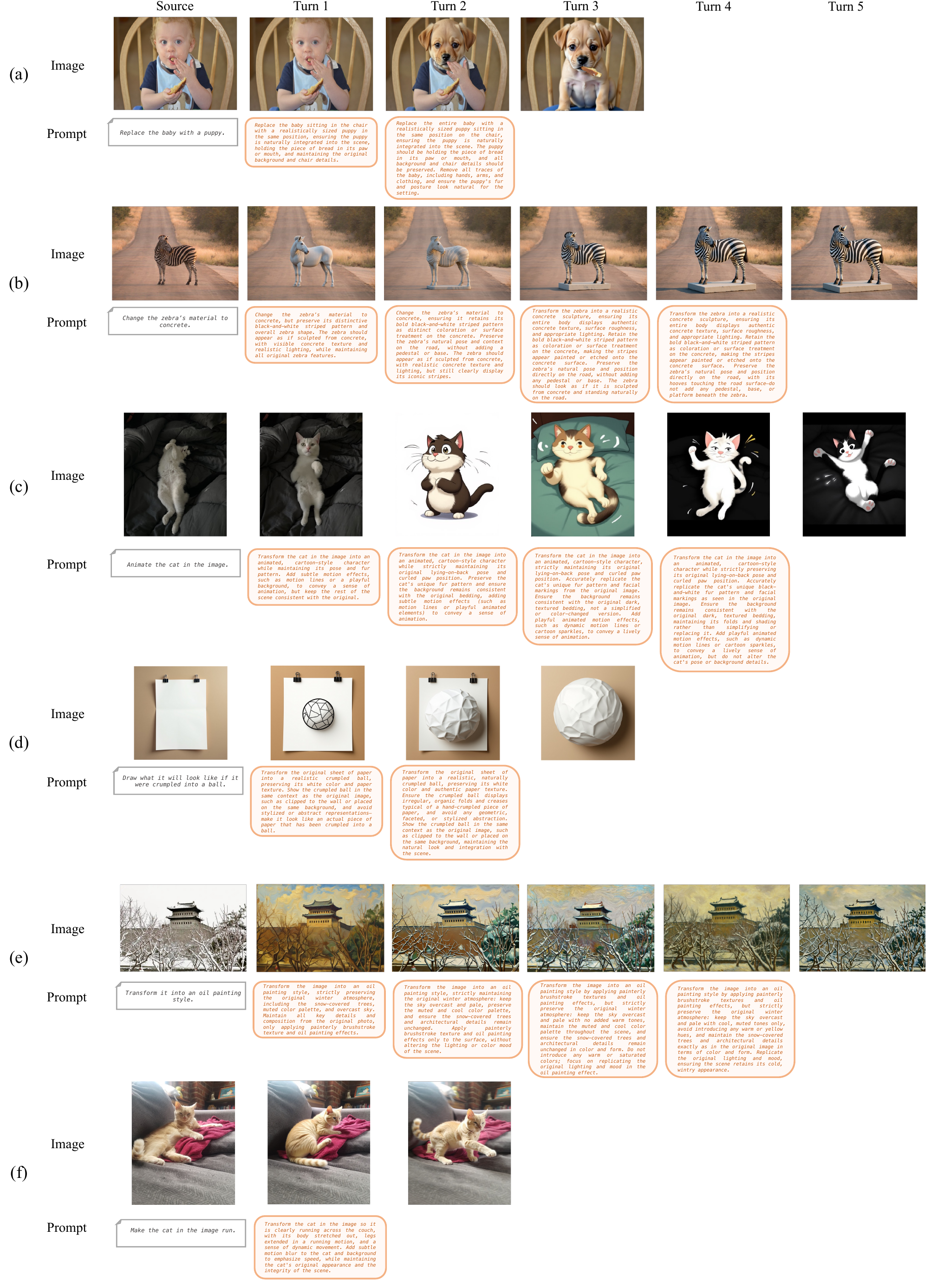} 
    \caption{
Qualitative visualizations of EditThinker paired with different editors.
Subfigures (a) and (b) show results with FLUX.1 Kontext [Dev], (c) and (d) use OmniGen2, and (e) and (f) use Qwen-Image-Edit.
}
    \label{fig:vis-all}

\end{figure}

\begin{figure}[p]
    \centering 
    \includegraphics[width=\linewidth,height=0.89\textheight,keepaspectratio]{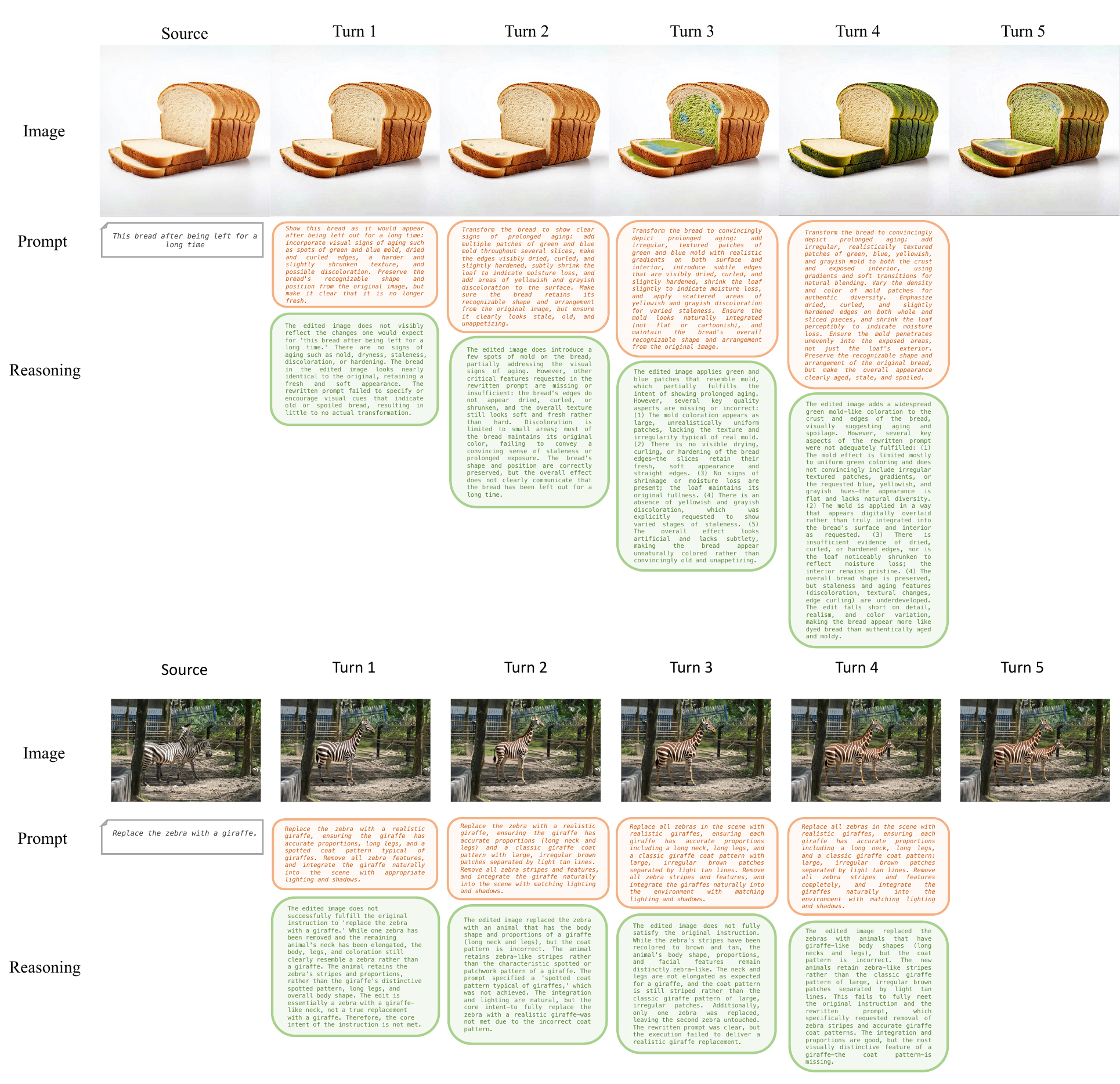} 
    \caption{Visualization of EditThinker’s reasoning traces and intermediate editing results when paired with FLUX.1 Kontext [Dev]. The figure illustrates how the Thinker evaluates the current output, identifies issues, and iteratively refines the instruction over multiple rounds.} 
    \label{fig:vis-reason}

\end{figure}
\clearpage
\setcounter{figure}{7}

\section{Qualitative Analysis}
\label{app:qualitative}
\label{sec:additional_qualitative}

\begin{figure}[p]
\centering
\includegraphics[pagebox=cropbox,width=\linewidth,trim=0bp 236bp 707bp 30bp,clip]{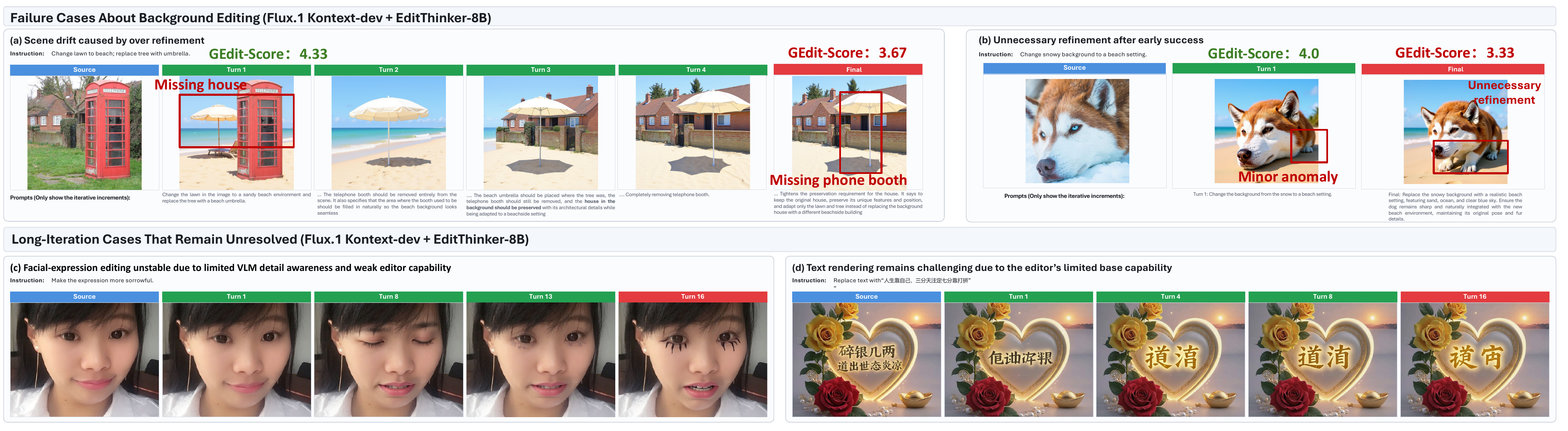}\par\medskip
\includegraphics[pagebox=cropbox,width=\linewidth,trim=1100bp 236bp 8bp 30bp,clip]{figs/failure-case.pdf}\par\medskip
\includegraphics[pagebox=cropbox,width=\linewidth,trim=0bp 8bp 903bp 289bp,clip]{figs/failure-case.pdf}\par\medskip
\includegraphics[pagebox=cropbox,width=\linewidth,trim=894bp 8bp 8bp 289bp,clip]{figs/failure-case.pdf}
\caption{Failure and unresolved cases with FLUX.1-Kontext-dev and EditThinker-8B. Original panels (a)--(d) are arranged vertically: (a) background editing changes unrequested scene content; (b) further refinement degrades an early result; (c) facial-expression editing and (d) text replacement remain unresolved through the displayed turn 16 in extended qualitative sequences. Scores and annotations are retained from the original examples.}
\label{fig:additional_failure_cases}
\end{figure}

\Cref{fig:additional_failure_cases} shows background-editing failures and long sequences with FLUX.1-Kontext-dev and EditThinker-8B. In panel (a), the request changes the lawn to a beach and replaces a tree with an umbrella, but later edits remove the phone booth; the displayed GEdit score falls from 4.33 to 3.67. In panel (b), further refinement of a dog-on-a-beach edit introduces additional changes, and the displayed score decreases from 4.0 to 3.33. These examples show that extra refinement can alter content beyond the requested background change.

Panels (c) and (d) show extended facial-expression and Chinese-text editing sequences through 16 editing rounds. These long-sequence visualizations examine persistent errors beyond the five-round benchmark horizon in Appendix~\ref{sec:additional_stopping}. The final facial-expression edit contains conspicuous eye-region artifacts, and the displayed Chinese text still does not match the requested replacement.

\begin{figure}[p]
\centering
\includegraphics[pagebox=cropbox,width=\linewidth,trim=0bp 9bp 1500bp 43bp,clip]{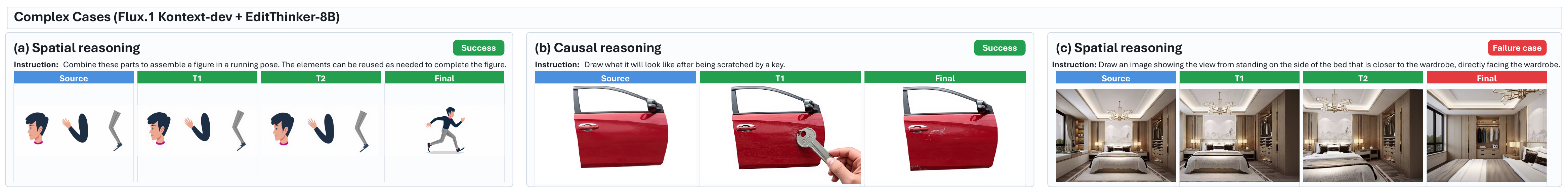}\par\medskip
\includegraphics[pagebox=cropbox,width=\linewidth,trim=747bp 9bp 757bp 43bp,clip]{figs/hard-case.pdf}\par\medskip
\includegraphics[pagebox=cropbox,width=\linewidth,trim=1492bp 9bp 0bp 43bp,clip]{figs/hard-case.pdf}
\caption{Challenging instructions with FLUX.1-Kontext-dev and EditThinker-8B. Original panels (a)--(c) are arranged vertically: assembling a running figure, depicting a scratched car door, and changing the bedroom viewpoint. The source examples mark the first two as successes and the third as a failure. Original instructions and intermediate results are preserved.}
\label{fig:additional_hard_cases}
\end{figure}

\Cref{fig:additional_hard_cases} contains two examples marked successful and one marked unsuccessful. Panel (a) assembles separate body parts into a running figure. Panel (b) changes the car-door image to show scratches, without retaining the key and hand that appear in the intermediate result. Panel (c) changes the bedroom viewpoint but remains labeled as failing the requested view from beside the bed toward the wardrobe.

\end{document}